\documentclass{article}

\usepackage{authblk}

\usepackage[utf8]{inputenc} 
\usepackage[T1]{fontenc}    
\usepackage{hyperref}       
\usepackage{url}            
\usepackage{booktabs}       
\usepackage{amsfonts}       
\usepackage{nicefrac}       
\usepackage{microtype}      
\usepackage{algorithm}
\usepackage{algorithmic}
\usepackage{amsmath}
\usepackage{times}
\usepackage{mathtools}
\usepackage[numbers]{natbib}
\usepackage{color}
\usepackage{siunitx}
\usepackage{multirow}
\usepackage{amsthm}
\usepackage{mathrsfs}
\usepackage{graphicx}
\usepackage{caption}
\usepackage{xspace}
\usepackage[export]{adjustbox}
\usepackage{float}
\usepackage{enumitem}
\usepackage{listings}
\usepackage{multirow}
\usepackage{bbm}
\usepackage{wrapfig}
\usepackage{amsthm}
\usepackage{soul} 
\usepackage{subcaption}
\usepackage{wrapfig}
\newtheorem{theorem}{Theorem}

\usepackage{tikz}
\usetikzlibrary{calc}
\tikzset{mynode/.style={draw,circle, minimum size = 0.7cm}}
\usetikzlibrary{positioning,shapes,arrows}
\usepackage{xcolor}
\usepackage{stmaryrd}

\definecolor{gblue}{RGB}{207,226,243}
\definecolor{gred}{RGB}{244,204,204}
\definecolor{gyellow}{RGB}{255,229,153}
\definecolor{gyellow2}{RGB}{252,229,205}
\definecolor{ggreen}{RGB}{217,234,211}
\definecolor{ggray}{RGB}{238,238,238} 
\definecolor{ggray2}{RGB}{81,84,87} 
\definecolor{gpurple}{RGB}{217,210,233} 

\usepackage{amsmath,amsfonts,bm}

\def\1{\bm{1}}

\def\rvu{{\mathbf{i}}}

\def\rvu{{\mathbf{u}}}

\def\rvx{{\mathbf{x}}}

\def\rvz{{\mathbf{z}}}

\def\vx{{\bm{x}}}

\DeclareMathAlphabet{\mathsfit}{\encodingdefault}{\sfdefault}{m}{sl}
\SetMathAlphabet{\mathsfit}{bold}{\encodingdefault}{\sfdefault}{bx}{n}

\def\gM{{\mathcal{M}}}

\def\gS{{\mathcal{S}}}

\def\gX{{\mathcal{X}}}

\def\gZ{{\mathcal{Z}}}

\def\sR{{\mathbb{R}}}

\def\sZ{{\mathbb{Z}}}

\newcommand{\R}{\mathbb{R}}

\newcommand{\KL}{D_{\mathrm{KL}}}

\usepackage{xcolor}
\hypersetup{
    colorlinks,
    anchorcolor={blue!50!black},
    linkcolor={blue!50!black},
    citecolor={blue!50!black},
    urlcolor={blue}
}

\usepackage{footnote}
\makesavenoteenv{tabular}

\newcommand{\supp}{\operatorname{supp}}

\usepackage[accepted]{icml2020}

\icmltitlerunning{Weakly-Supervised Disentanglement Without Compromises}

\begin{document}

\twocolumn[
\icmltitle{Weakly-Supervised Disentanglement Without Compromises}

\icmlsetsymbol{equal}{*}

\begin{icmlauthorlist}
\icmlauthor{Francesco Locatello}{eth,mpi}
\icmlauthor{Ben Poole}{google}
\icmlauthor{Gunnar R\"atsch}{eth}\\
\icmlauthor{Bernhard Sch\"olkopf}{mpi}
\icmlauthor{Olivier Bachem}{google}
\icmlauthor{Michael Tschannen}{google}
\end{icmlauthorlist}

\icmlaffiliation{eth}{Department of Computer Science, ETH Zurich}
\icmlaffiliation{mpi}{Max Planck Institute for Intelligent Systems}
\icmlaffiliation{google}{Google Research, Brain Team}

\icmlcorrespondingauthor{}{francesco.locatello@inf.ethz.ch}
\icmlkeywords{Disentanglement}

\vskip 0.3in
]

\printAffiliationsAndNotice{}  

\begin{abstract}
\looseness=-1Intelligent agents should be able to learn useful representations by observing changes in their environment. We model such observations as pairs of non-i.i.d. images sharing at least one of the underlying factors of variation.
First, we theoretically show that only knowing \textit{how many} factors have changed, but not which ones, is sufficient to learn disentangled representations. Second, we provide practical algorithms that learn disentangled representations from pairs of images without requiring annotation of groups, individual factors, or the number of factors that have changed.
Third, we perform a large-scale empirical study and show that such pairs of observations are sufficient to reliably learn disentangled representations on several benchmark data sets. Finally, we evaluate our learned representations and find that they are \emph{simultaneously} useful on a diverse suite of tasks, including generalization under covariate shifts, fairness, and abstract reasoning. Overall, our results demonstrate that weak supervision enables learning of useful disentangled representations in realistic scenarios.
\end{abstract}

\section{Introduction}
\looseness=-1A recent line of work argued that representations which are \textit{disentangled} offer useful properties such as interpretability~\cite{adel2018discovering,bengio2013representation,higgins2016beta}, predictive performance~\cite{locatello2018challenging,locatello2019disentangling}, reduced sample complexity on abstract reasoning tasks~\cite{van2019disentangled}, and fairness~\cite{locatello2019fairness,creager2019flexibly}.
The key underlying assumption is that high-dimensional observations $\rvx$ (such as images or videos) are in fact a manifestation of a low-dimensional set of independent ground-truth factors of variation $\rvz$~\cite{locatello2018challenging,bengio2013representation,tschannen2018recent}. The goal of disentangled representation learning is to learn a function $r(\rvx)$ mapping the observations to a low-dimensional vector that contains all the information about each factor of variation, with each coordinate (or a subset of coordinates) containing information about only one factor.
Unfortunately,~\citet{locatello2018challenging} showed that the unsupervised learning of disentangled representations is theoretically impossible from i.i.d. observations without inductive biases. In practice, they observed that unsupervised models exhibit significant variance depending on hyperparameters and random seed, making their training somewhat unreliable.

\begin{figure}
\hspace{-0.3cm}\makebox[0.42\textwidth][c]{%
    \centering
    \definecolor{mygreen}{RGB}{30,142,62}
	\definecolor{myred}{RGB}{217,48,37}
	\definecolor{myw}{RGB}{255,255,255}
	\scalebox{0.7}{
	\centering
	\begin{minipage}{.28\textwidth}
	\begin{tikzpicture}[scale=1.2]
		\tikzset{
			myarrow/.style={->, >=latex', shorten >=1pt, line width=0.4mm},
		}
		\tikzset{
			mybigredarrow/.style={dashed, dash pattern={on 20pt off 10pt}, >=latex', shorten >=3mm, shorten <=3mm, line width=2mm, draw=myred},
		}
		\tikzset{
			mybiggreenarrow/.style={->, >=latex', shorten >=3mm, shorten <=3mm, line width=2mm, draw=mygreen},
		}
		\node[text width=.8cm,align=center,minimum size=2.6em,draw,thick,circle, fill=myw] (z) at (-5.25,1.5) {$\rvz$};
		\node[text width=.8cm,align=center,minimum size=2.6em,draw,thick,circle, fill=myw] (zt) at (-5.25,0) {$\tilde\rvz$};
		\node[text width=.8cm,align=center,minimum size=2.6em,draw,thick,circle, fill=myw] (s) at (-5.25,-1.5) {$S$};
		
		\node[text width=.8cm,align=center,minimum size=2.6em,draw,thick,circle, fill=ggray] (x1) at (-2.75,1) {$\rvx_1$};
		
		\node[text width=.8cm,align=center,minimum size=2.6em,draw,thick,circle, fill=ggray] (x2) at (-2.75,-1) {$\rvx_2$};
		\draw[myarrow] (z) -- (x1);
		\draw[myarrow] (z) -- (x2);
		\draw[myarrow] (zt) -- (x2);
		\draw[myarrow] (s) -- (x2);

	\end{tikzpicture}
	\end{minipage}%
	\begin{minipage}{.2\textwidth}
	\begin{tikzpicture}
		\tikzset{
			myarrow/.style={->, >=latex', shorten >=1pt, line width=0.4mm},
		}
		\tikzset{
			mybigredarrow/.style={dashed, dash pattern={on 20pt off 10pt}, >=latex', shorten >=3mm, shorten <=3mm, line width=2mm, draw=myred},
		}
		\tikzset{
			mybiggreenarrow/.style={->, >=latex', shorten >=3mm, shorten <=3mm, line width=2mm, draw=mygreen},
		}
		\node[fill=myw] (z1) at (-5.25,1.7) {
		$\begin{bmatrix}
           2 \\
           1 \\
           1 \\
           3 \\
           \color{red}\textbf{1} \\
           \color{red}\textbf{1} \\
           1 \\
         \end{bmatrix}
		$};

		\node[] (z) at (-5.75,1.7) {
		$\rvz\left\{
            \begin{array}{ c }
            \\
            \\
            \\
            \\
            \\
            \\
            \\
            \end{array}
        \right.
		$};
		
		\node[] (zt) at (-5.75,-2.34) {
		$\tilde\rvz\left\{
            \begin{array}{ c }
            \\
            \\
            \end{array}
        \right.
		$};

		\node[fill=myw] (z2) at (-5.25,-1.7) {$\begin{bmatrix}
           2 \\
           1 \\
           1 \\
           3 \\
           \color{red}\textbf{8} \\
           \color{red}\textbf{2} \\
           1 \\
         \end{bmatrix}
		$};

		\node[inner sep=0pt] (x1) at (-2.,1.25) {\includegraphics[width=.7 \textwidth]{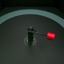}};
		
		\node[inner sep=0pt] (x2) at (-2.,-1.25) {\includegraphics[width=.7\textwidth]{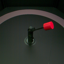}};
		
		\draw[myarrow] (z1) -- (x1);
		\draw[myarrow] (z2) -- (x2);

	\end{tikzpicture}
	\end{minipage}
	}}
	
    \caption{(\textbf{left}) The proposed generative model. We observe pairs of observations $(\rvx_1, \rvx_2)$ sharing a random subset $S$ of latent factors: $\rvx_1$ is generated by $\rvz$; $\rvx_2$ is generated by combining the subset $S$ of $\rvz$ and resampling the remaining entries (modeled by $\tilde \rvz$). (\textbf{right}) Real-world example of the model: A pair of images from MPI3D~\cite{gondal2019transfer} where all factors are shared except the first degree of freedom and the background color (red values). This corresponds to a setting where few factors in a causal generative model change, which, by the {\em independent causal mechanisms} principle, leaves the others invariant \cite{scholkopf2012causal}.}
    \label{fig:disent_cause}
    \vspace{-0.5cm}
\end{figure}

\looseness=-1On the other hand, many data modalities are \textit{not} observed as i.i.d. samples from a distribution~\cite{dayan1993improving,storck1995reinforcement,hochreiter1999feature,bengio2013representation,PetJanSch17,thomas2018disentangling,1911.10500}. Changes in natural environments, which typically correspond to changes of only a few underlying factors of variation, provide a weak supervision signal for representation learning algorithms~\cite{foldiak1991learning,schmidt2007learning,bengio2017consciousness,bengio2019meta}. 
State-of-the-art weakly-supervised disentanglement methods~\cite{bouchacourt2017multi,gvae2019,shu2020weakly} assume that observations belong to annotated groups where two things are known at training time: (i) the relation between images in the same group, and (ii) the group each image belongs to. \citet{bouchacourt2017multi,gvae2019} consider groups of observations differing in precisely one of the underlying factors. An example of such a group are images of a given object with a fixed orientation, in a fixed scene, but of varying color. \citet{shu2020weakly} generalized this notion to other relations (e.g., single shared factor, ranking information).
In general, precise knowledge of the groups and their structure may require either explicit human labeling or at least strongly controlled acquisition of the observations. As a motivating example, consider the video feedback of a robotic arm. In two temporally close frames, both the manipulated objects and the arm may have changed their position, the objects themselves may be different, or the lighting conditions may have changed due to failures.

In this paper, we consider learning disentangled representations from pairs of observations which differ by a few factors of variation~\cite{bengio2017consciousness,schmidt2007learning,bengio2019meta} as in Figure~\ref{fig:disent_cause}.
Unlike previous work on weakly-supervised disentanglement, we consider the realistic and broadly applicable setting where we observe pairs of images and have no additional annotations: It is unknown \textit{which} and \textit{how many} factors of variation have changed. In other words, we do not know which group each pair belongs to, and what is the precise relation between the two images.
The only condition we require is that the two observations are different and that the change in the factors is not dense. The key contributions of this paper are:
\begin{itemize}[itemsep=0.5pt,topsep=1pt, leftmargin=10pt]
    \item We present simple adaptive group-based disentanglement methods which do not require annotations of the groups, as opposed to~\cite{bouchacourt2017multi,gvae2019,shu2020weakly}.
    Our approach is readily applicable to a variety of settings where groups of non-i.i.d. observations are available with no additional annotations.
    \item  We theoretically show that identifiability is possible from non-i.i.d. pairs of observations under weak assumptions. Our proof motivates the setup we consider, which is identifiable as opposed to the standard one, which was proven to be non-identifiable~\citep{locatello2018challenging}. Further, we use theoretical arguments to inform the design of our algorithms, recover existing group-based VAE methods~\cite{bouchacourt2017multi,gvae2019} as special cases, and relax their impractical assumptions.
    \item  \looseness=-1We perform a large-scale reproducible experimental study training over \num{15000} disentanglement models and over one million downstream classifiers\footnote{Our experiments required $\sim\!5.85$ GPU years (NVIDIA P100).} on five different data sets, one of which consisting of real images of a robotic platform~\cite{gondal2019transfer}. 
    \item We demonstrate that one can reliably learn disentangled representations with weak supervision only, \emph{without relying on supervised disentanglement metrics for model selection}, as done in previous works. 
    Further, we show that these representations are useful on a diverse suite of downstream tasks, including a novel experiment targeting strong generalization under covariate shifts, fairness~\cite{locatello2019fairness} and abstract visual reasoning~\cite{van2019disentangled}.
\end{itemize}

\section{Related work}\label{sec:other_rel_work}
\looseness=-1Recovering independent components of the data generating process is a well-studied problem in machine learning. It has roots in the independent component analysis (ICA) literature, where the goal is to unmix independent non-Gaussian sources of a $d$-dimensional signal~\citep{comon1994independent}.
Crucially, identifiability is not possible in the nonlinear case from i.i.d. observations~\citep{hyvarinen1999nonlinear}. Recently, the ICA community has considered weak forms of supervision such as temporal consistency~\cite{hyvarinen2016unsupervised,hyvarinen2017nonlinear}, auxiliary supervised information~\cite{hyvarinen2018nonlinear,khemakhem2019variational}, and multiple views~\cite{gresele2019incomplete}. A parallel thread of work has studied distribution shifts by identifying changes in causal generative factors \citep{zhang_multi-source_2015,Zhangetal17,HuaZhaZhaSanGlySch17}, which is linked to a causal view of disentanglement \citep{suter2018interventional,1911.10500}.

\looseness=-1On the other hand, more applied machine learning approaches have experienced the opposite shift. Initially, the community focused on more or less explicit and task dependent supervision~\citep{reed2014learning,yang2015weakly,kulkarni2015deep,cheung2014discovering,mathieu2016disentangling,narayanaswamy2017learning}.
For example, a number of works rely on known relations between the factors of variation~\citep{karaletsos2015bayesian,whitney2016understanding,fraccaro2017disentangled,denton2017unsupervised,hsu2017unsupervised,yingzhen2018disentangled,locatello2018clustering,ridgeway2018learning,chen2019weakly} and disentangling motion and pose from content~\citep{hsieh2018learning,fortuin2018deep,deng2017factorized,goroshin2015learning}.

\looseness=-1Recently, there has been a renewed interest in the unsupervised learning of disentangled representations~\citep{higgins2016beta,burgess2018understanding,kim2018disentangling,chen2018isolating,kumar2017variational} along with quantitative evaluation~\citep{kim2018disentangling,eastwood2018framework,kumar2017variational,ridgeway2018learning,duan2019heuristic}. After the theoretical impossibility result of~\citet{locatello2018challenging}, the focus shifted back to semi-supervised~\cite{locatello2019disentangling,sorrenson2020disentanglement,khemakhem2019variational} and weakly-supervised approaches~\cite{bouchacourt2017multi,gvae2019,shu2020weakly}. 

\section{Generative models} \label{sec:model-ident-algo}
We first describe the generative model commonly used in the disentanglement literature, and then turn to the weakly-supervised model used in this paper.

\textbf{Unsupervised generative model }
First, a $\rvz$ is drawn from a set of independent ground-truth factors of variation $p(\rvz) = \prod_i p(\rvz_i)$. Second, the observations are obtained as draws from $p(\rvx|\rvz)$. The factors of variation $\rvz_i$ do not need to be one-dimensional but we assume so to simplify the notation.

\textbf{Disentangled representations} The goal of disentanglement learning is to learn a mapping $r(\rvx)$ where the effect of the different factors of variation is axis-aligned with different coordinates. More precisely, each factor of variation $z_i$ is associated with exactly one coordinate (or group of coordinates) of $r(\rvx)$ and vice-versa (and the groups are non-overlapping). As a result, varying one factor of variation and keeping the others fixed results in a variation of exactly one coordinate (group of coordinates) of $r(\rvx)$. \citet{locatello2018challenging} theoretically showed that learning such a mapping $r$ is theoretically impossible without inductive biases or some other, possibly weak, form of supervision. 

\textbf{Weakly-supervised generative model} We study learning of disentangled image representations from paired observations, for which some (but not all) factors of variation have the same value. This can be modeled as sampling two images from the causal generative model with an intervention~\cite{PetJanSch17} on a random subset of the factors of variation. Our goal is to use the additional information given by the pair (as opposed to a single image) to learn a disentangled image representations. 
We generally do not assume knowledge of which or how many factors are shared, i.e., we do not require controlled acquisition of the observations. This observation model applies to many practical scenarios. For example, we may want to learn a disentangled representation of a robot arm observed through a camera: In two temporally close frames some joint angles will likely have changed, but others will have remained constant. Other factors of variation may also change independently of the actions of the robot. An example can be seen in Figure~\ref{fig:disent_cause} (right) where the first degree of freedom of the arm and the color of the background changed. More generally this observation model applies to many natural scenes with moving objects~\citep{foldiak1991learning}. 
\newcommand{\dx}{\mathrm{d}}
\looseness=-1More formally, we consider the following generative model.
For simplicity of exposition, we assume that the number of factors $k$ in which the two observations differ is constant (we present a strategy to deal with varying $k$ in Section~\ref{sec:algorithms}). 
The generative model is given by 
\begin{align}
    p(\rvz) &= \prod_{i=1}^d p(z_i), \quad p(\tilde \rvz) = \prod_{i=1}^{k} p(\tilde z_i), \quad S \sim p(S) \label{eq:generative-model-1}\\
    \rvx_1 &= g^\star(\rvz), \qquad 
    \rvx_2 = g^\star(f(\rvz, \tilde \rvz, S)), \label{eq:generative-model-4}
\end{align}
where $S$ is the subset of shared indices of size $d-k$ sampled from a distribution $p(S)$ over the set 
$\gS = \{ S \subset [d] \colon |S|=d-k \}$, and the $p(z_i)$ and $p(\tilde z_j)$ are all identical. 
The generative mechanism is modeled using a function
$g^\star \colon \gZ \to \gX$, with $\gZ = \supp(\rvz) \subseteq \sR^d$ and $\gX \subset \sR^m$, which maps the latent variable to observations of dimension $m$, typically $m \gg d$. To make the relation between $\rvx_1$ and $\rvx_2$ explicit, we use a function $f$ obeying
\begin{equation*}
    f(\rvz, \tilde \rvz, S)_S = \rvz_S \qquad \text{and} \qquad f(\rvz, \tilde \rvz, S)_{\bar S} = \tilde \rvz 
\end{equation*}
\looseness=-1with $\bar S = [d] \backslash S$. Intuitively, to generate $\rvx_2$, $f$ selects entries from $\rvz$ with index in $S$ and substitutes the remaining factors with $\tilde \rvz$, thus ensuring that the factors indexed by $S$ are shared in the two observations. 
The generative model \eqref{eq:generative-model-1}--\eqref{eq:generative-model-4} does not model additive noise; we assume that noise is explicitly modeled as a latent variable and its effect is manifested through $g^\star$ as done by~\cite{bengio2013representation,locatello2018challenging,higgins2018towards,higgins2016beta,suter2018interventional,reed2015deep,lecun2004learning,kim2018disentangling,gondal2019transfer}.
For simplicity, we consider the case where groups consisting of two observations (pairs), but extensions to more than two observations are possible~\cite{gresele2019incomplete}. 

\section{Identifiability and algorithms}
First, we show that, as opposed to the unsupervised case~\cite{locatello2018challenging}, the generative model \eqref{eq:generative-model-1}--\eqref{eq:generative-model-4} is identifiable under weak additional assumptions. Note that the joint distribution of all random variables factorizes as
\begin{equation}
    p(\rvx_1, \rvx_2, \rvz, \tilde \rvz, S) = p(\rvx_1 | \rvz)p(\rvx_2 | f(\rvz, \tilde \rvz, S)) p(\rvz) p(\tilde \rvz) p(S) \label{eq:full-joint}
\end{equation}
where the likelihood terms have the same distribution, i.e., $p(\rvx_1 | \bar \rvz) = p(\rvx_2 | \bar \rvz), \forall \bar \rvz \in \supp(p(\rvz))$. We show that to learn a disentangled generative model of the data $p(\rvx_1, \rvx_2)$ it is therefore sufficient to recover a factorized latent distribution with factors $p(\hat z_i)= p(\hat{\tilde{z}}_j)$, a corresponding likelihood $q(\rvx_1|\cdot)=q(\rvx_2|\cdot)$, as well as a distribution $p(\hat S)$ over $\gS$, which together satisfy the constraints of the true generative model \eqref{eq:generative-model-1}--\eqref{eq:generative-model-4} and match the true $p(\rvx_1, \rvx_2)$ after marginalization over $\hat \rvz,\hat{\tilde{\rvz}}, \hat S$ when substituted into \eqref{eq:full-joint}.

\begin{theorem} \label{thm:identifiability}
Consider the generative model \eqref{eq:generative-model-1}--\eqref{eq:generative-model-4}. Further assume that $p(z_i) = p(\tilde z_i)$ are continuous distributions, $p(S)$ is a distribution over $\gS$ s.t. for $S, S' \sim p(S)$ we have $P(S \cap S' = \{i\}) > 0, \forall i \in [d]$. Let $g^\star \colon \gZ \to \gX$ in \eqref{eq:generative-model-4} be smooth and invertible on $\gX$ with smooth inverse (i.e., a diffeomorphism). Given unlimited data from $p(\rvx_1, \rvx_2)$ and the true (fixed) $k$, consider all tuples $(p(\hat z_i), q(\rvx_1|\hat \rvz), p(\hat S))$ obeying these assumptions and matching $p(\rvx_1, \rvx_2)$ after marginalization over $\hat \rvz,\hat{\tilde{\rvz}}, \hat S$ when substituted in \eqref{eq:full-joint}. Then, the posteriors $q(\hat \rvz | \rvx_1) = q(\rvx_1|\hat \rvz)p(\hat\rvz)/p(\rvx_1)$ are disentangled in the sense that the aggregate posteriors $q(\hat \rvz) = \int q(\hat \rvz|\rvx_1) p(\rvx_1) \dx \rvx_1 = \iint q(\hat \rvz|\rvx_1) p(\rvx_1|\rvz) p(\rvz)\dx \rvz \dx \rvx_1$ are coordinate-wise reparameterizations of the ground-truth prior $p(\rvz)$ up to a permutation of the indices of $\rvz$.
\end{theorem}

\looseness=-1\textbf{Discussion} 
Under the assumptions of this theorem, we established that all generative models that match the true marginal over the observations $p(\rvx_1, \rvx_2)$ must be disentangled. Therefore, constrained distribution matching is sufficient to learn disentangled representations. Formally, the aggregate posterior $q(\hat \rvz)$ is a coordinate-wise reparameterization of the true distribution of the factors of variation (up to index permutations). In other words, there exists a one-to-one mapping between every entry of $\rvz$ and a unique matching entry of $\hat \rvz$, and thus a change in a single coordinate of $\rvz$ implies a change in a single matching coordinate of $\hat \rvz$~\cite{bengio2013representation}. Changing the observation model from single i.i.d. observations to non-i.i.d. pairs of observations generated according to the generative model \eqref{eq:generative-model-1}--\eqref{eq:generative-model-4} allows us to bypass the non-identifiability result of~\cite{locatello2018challenging}.
Our result requires strictly weaker assumptions than the result of~\citet{shu2020weakly} as we do not require group annotations, but only knowledge of $k$.  As we shall see in Section~\ref{sec:algorithms}, $k$ can be cheaply and reliably estimated from data at run-time. 
Although the weak assumptions of Theorem~\ref{thm:identifiability} may not be satisfied in practice, we will show that the proof can inform practical algorithm design.

\subsection{Practical adaptive algorithms} \label{sec:algorithms}
We conceive two $\beta$-VAE \cite{higgins2016beta} variants tailored to the weakly-supervised generative model \eqref{eq:generative-model-1}--\eqref{eq:generative-model-4} and a selection heuristic to deal with unknown and random $k$. We will see that these simple models can very reliably learn disentangled representations. 

\looseness=-1The key differences between theory and practice are that: (i) we use the ELBO and amortized variational inference for distribution matching (the true and learned distributions will not exactly match after training), (ii) we have access to a finite number of data only, and (iii) the theory assumes known, fixed $k$, but $k$ might be unknown and random. 

\looseness=-1\textbf{Enforcing the structural constraints }
Here we present a simple structure for the variational family that allows us to tractably perform approximate inference on the weakly-supervised generative model.
First note that the alignment constraints imposed by the generative model (see \eqref{eq:gconstraint1v2} and \eqref{eq:gconstraint2v2} evaluated for $g = g^\star$ in Appendix~\ref{sec:proof}) imply for the true posterior
\begin{align}
p(z_i | \rvx_1) &= p(z_i | \rvx_2) \quad \forall i \in S, \label{eq:pconstraint1}\\
p(z_i | \rvx_1) &\neq p(z_i | \rvx_2) \quad \forall i \in \bar S, \label{eq:pconstraint2}
\end{align}
(with probability $1$)
and we want to enforce these constraints on the approximate posterior $q_\phi(\hat\rvz | \rvx)$ of our learned model. However, the set $S$ is unknown. To obtain an estimate $\hat S$ of $S$ we therefore choose for every pair $(\rvx_1, \rvx_2)$ the $d-k$ coordinates with the smallest $\KL(q_\phi(\hat z_i | \rvx_1)||q_\phi(\hat z_i | \rvx_2))$. To impose the constraint \eqref{eq:pconstraint1} we then replace each shared coordinate with some average $a$ of the two posteriors
\begin{align*}
    \tilde q_\phi(\hat z_i | \rvx_1) &= a(q_\phi(\hat z_i | \rvx_1), q_\phi(\hat z_i | \rvx_2)) \quad &\forall i \in \hat S, \\
    \tilde q_\phi(\hat z_i | \rvx_1) &= q_\phi(\hat z_i | \rvx_1) &\text{else,}
\end{align*}
and obtain $\tilde q_\phi(\rvz_i | \rvx_2)$ in analogous manner. As we later simply use the averaging strategies of the Group-VAE (GVAE)~\citep{gvae2019} and the Multi Level-VAE (ML-VAE)~\citep{bouchacourt2017multi}, we term variants of our approach which infers the groups and their properties adaptively \textit{Adaptive-Group-VAE} (Ada-GVAE) and \textit{Adaptive-ML-VAE} (Ada-ML-VAE), depending on the choice of the averaging function $a$. We then optimize the following variant of the $\beta$-VAE objective 
\begin{align}
    \max_{\phi, \theta} \mathbb{E}_{(\rvx_1, \rvx_2)}&\mathbb{E}_{\tilde q_\phi(\hat\rvz | \rvx_1)} \log(p_\theta(\rvx_1|\hat\rvz)) \nonumber \\
    &+ \mathbb{E}_{\tilde q_\phi(\hat\rvz | \rvx_2)} \log(p_\theta(\rvx_2|\hat\rvz)) \nonumber \\
    &- \beta D_{KL} \left(\tilde q_\phi(\hat\rvz || \rvx_1)|p(\hat\rvz)\right) \nonumber \\
    &- \beta D_{KL} \left(\tilde q_\phi(\hat\rvz || \rvx_2)|p(\hat\rvz)\right), \label{eq:mod-elbo}
\end{align}
where $\beta \geq 1$ \cite{higgins2016beta}. The advantage of this averaging-based implementation of \eqref{eq:pconstraint1}, over implementing it, for instance, via a $\KL$-term that encourages the distributions of the shared coordinates $\hat S$ to be similar, is that averaging imposes a hard constraint in the sense that $q_\phi(\hat\rvz|\rvx_1)$ and $q_\phi(\hat\rvz|\rvx_2)$ can jointly encode only one value per shared coordinate. This in turn implicitly enforces the constraint \eqref{eq:pconstraint2} as the non-shared dimensions need to be efficiently used to encode the non-shared factors of $\rvx_1$ and $\rvx_2$. 

\looseness=-1We emphasize that the objective \eqref{eq:mod-elbo} is a simple modification of the $\beta$-VAE objective and is very easy to implement. Finally, we remark that invoking Theorem~4 of~\cite{khemakhem2019variational}, we achieve consistency under maximum likelihood estimation up to the equivalence class in our Theorem~\ref{thm:identifiability}, for $\beta=1$ and in the limit of infinite data and capacity. 

\textbf{Inferring $k$} In the (practical) scenario where $k$ is unknown, we use the threshold
$$
    \tau = \frac{1}{2}(\max_i \delta_i + \min_i \delta_i),
$$
where $\delta_i = \KL(q_\phi(\hat z_i | \rvx_1)||q_\phi(\hat z_i | \rvx_2))$, and average the coordinates with $\delta_i < \tau$. This heuristic is inspired by the ``elbow method''~\cite{ketchen1996application} for model selection in $k$-means clustering and $k$-singular value decomposition and we found it to work surprisingly well in practice (see the experiments in Section~\ref{sec:results}). This estimate relies on the assumption that not all factors have changed. All our adaptive methods use this heuristic.
Although a formal recovery argument cannot be made for arbitrary data sets, inductive biases may limit the impact of an approximate $k$ in practice. We further remark that this heuristic always yields the correct $k$ if the encoder is disentangled.

\looseness=-1\textbf{Relation to prior work} Closely related to the proposed objective \eqref{eq:mod-elbo} the GVAE of~\citet{gvae2019} and the ML-VAE of~\citet{bouchacourt2017multi} assume $S$ is known and implement $a$ using different averaging choices. 
Both assume Gaussian approximate posteriors where $\mu_j, \Sigma_j$ are the mean and variance of $q(\hat \rvz_{ S}|\rvx_j)$ and $\mu, \Sigma$ are the mean and variance, of $\tilde q(\hat \rvz_{ S}|\rvx_j)$.
For the coordinates in $S$, the GVAE uses a simple arithmetic mean ($\mu = \frac12(\mu_1 + \mu_2)$ and $\Sigma = \frac12(\Sigma_1 + \Sigma_2)$) and the ML-VAE takes the product of the encoder distributions, with $\mu, \Sigma$ taking the form:
\begin{align*}
    \Sigma^{-1} = \Sigma^{-1}_1 + \Sigma^{-1}_2, \quad \mu^T = (\mu_1^T\Sigma_1^{-1} + \mu_2^T\Sigma_2^{-1})\Sigma.
\end{align*}
Our approach critically differs in the sense that $S$ is not known and needs to be estimated for every pair of images.

\looseness=-1Recent work combines non-linear ICA with disentanglement~\cite{khemakhem2019variational,sorrenson2020disentanglement}. 
Critically, these approaches are based on the setup of~\citet{hyvarinen2018nonlinear} which requires access to label information $\rvu$ such that $p(\rvz|\rvu)$ factorizes as $\prod_i p(z_i|\rvu)$. In contrast, we base our work on the setup of~\citet{gresele2019incomplete}, which only assumes access to two \textit{sufficiently distinct views} of the latent variable. \citet{shu2020weakly} train the same type of generative models over paired data but use a GAN objective where inference is not required. However, they require known and fixed $k$ as well as annotations of which factors change in each pair.

\section{Experimental results}\label{sec:results}

\looseness=-1\textbf{Experimental setup} We consider the setup of~\citet{locatello2018challenging}. We use the five data sets where the observations are generated as deterministic functions of the factors of variation: \textit{dSprites}~\cite{higgins2016beta}, \textit{Cars3D}~\cite{reed2015deep}, \textit{SmallNORB}~\cite{lecun2004learning}, \textit{Shapes3D}~\cite{kim2018disentangling}, and the real-world robotics data set \textit{MPI3D}~\cite{gondal2019transfer}.
\looseness=-1Our unsupervised baselines correspond to a cohort of \num{9000} unsupervised models ($\beta$-VAE~\citep{higgins2016beta}, AnnealedVAE~\citep{burgess2018understanding}, Factor-VAE~\citep{kim2018disentangling}, $\beta$-TCVAE~\citep{chen2018isolating}, DIP-VAE-I and II~\citep{kumar2017variational}), each with the same six hyperparameters from~\citet{locatello2018challenging} and 50 random seeds. 

\looseness=-1To create data sets with weak supervision from the existing disentanglement data sets, we first sample from the discrete $\rvz$ according to the ground-truth generative model~\eqref{eq:generative-model-1}--\eqref{eq:generative-model-4}. Then, we sample either one factor (corresponding to sparse changes) or $k$ factors of variation (to allow potentially denser changes) that may not be shared by the two images and re-sample those coordinates to obtain $\tilde \rvz$.  This ensures that each image pair differs in at most $k$ factors of variation (although changes are typically sparse and some pairs may be identical). For $k$ we consider the range from $1$ to $d-1$. This last setting corresponds to the case where all but one factor of variation are re-sampled. We study both the case where $k$ is constant across all pairs in the data set and where $k$ is sampled uniformly in the range $[d-1]$ for every training pair ($k=\texttt{Rnd}$ in the following). Unless specified otherwise, we aggregate the results for all values of $k$.

For each data set, we train four weakly-supervised methods: Our adaptive and vanilla (group-supervision) variants of GVAE~\cite{gvae2019} and ML-VAE~\cite{bouchacourt2017multi}. For each approach we consider six values for the regularization strength 
and 10 random seeds, training a total of \num{6000} weakly-supervised models. We perform model selection using the weakly-supervised reconstruction loss (i.e., the sum of the first two terms in \eqref{eq:mod-elbo})\footnote{In Figure~\ref{fig:rank_corr} in the appendix, we show that the training loss and the ELBO correlate similarly with disentanglement.}. We stress that we \textit{do not require labels for model selection}.

To evaluate the representations, we consider the disentanglement metrics in~\citet{locatello2018challenging}:  \emph{BetaVAE} score~\cite{higgins2016beta}, \emph{FactorVAE} score~\cite{kim2018disentangling}, \emph{Mutual Information Gap (MIG)}~\cite{chen2018isolating}, \emph{Modularity}~\cite{ridgeway2018learning}, \emph{DCI Disentanglement}~\cite{eastwood2018framework} and \emph{SAP score}~\cite{kumar2017variational}. To directly compare the disentanglement produced by different methods, we report the DCI Disentanglement~\cite{eastwood2018framework} in the main text and defer the plots with the other scores to the appendix as the same conclusions can be drawn based on these metrics. 
Appendix~\ref{sec:details} contains full implementation details.

\subsection{Is weak supervision enough for disentanglement?}\label{sec:results:learning_weak_sup}
\begin{figure*}[t]
    \centering
    \includegraphics[width=0.9\linewidth]{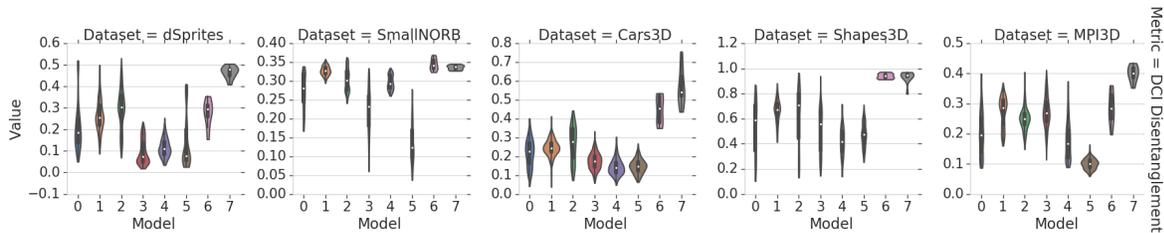}
    \vspace{-4mm}
    \caption{Our adaptive variants of the group-based disentanglement methods (models 6 and 7) significantly and consistently outperform unsupervised methods. In particular, the Ada-GVAE consistently yields the same or better performance than the Ada-ML-VAE. In this experiment, we consider the case where the number of shared factors of variation is random and different for every pair with high probability ($k=\texttt{Rnd}$).
    Legend: 0=$\beta$-VAE, 1=FactorVAE, 2=$\beta$-TCVAE, 3=DIP-VAE-I, 4=DIP-VAE-II, 5=AnnealedVAE, 6=Ada-ML-VAE, 7=Ada-GVAE}
    \label{fig:selection}
    \vspace{-3mm}
\end{figure*}

\looseness=-1In Figure~\ref{fig:selection}, we compare the performance of the weakly-supervised methods with $k=\texttt{Rnd}$ against the unsupervised methods. 
Unlike in unsupervised disentanglement with $\beta$-VAEs where $\beta\gg 1$ is common, we find  $\beta=1$ (the ELBO) performs best in most cases. We clearly observe that weakly-supervised models outperform the unsupervised ones.
In Figure~\ref{fig:selection_app} in the appendix, we further observe that they are competitive even if we allow fully supervised model selection on the unsupervised models. 
The Ada-GVAE performs similarly to the Ada-ML-VAE. For this reason, we focus the following analysis on the Ada-GVAE, and include Ada-ML-VAE results in Appendix~\ref{sec:results:app}.

\looseness=-1\textbf{Summary} With weak supervision, we reliably learn disentangled representations that outperform unsupervised ones. Our representations are competitive even if we perform fully supervised model selection on the unsupervised models.

\subsection{Are our methods adaptive to different values of $k$?}\label{sec:results_sup} 
\begin{figure}[t]
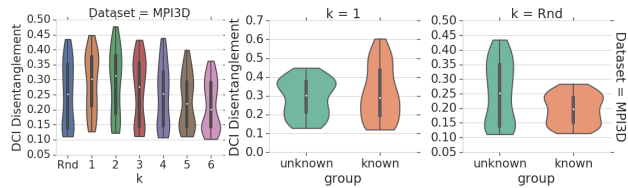

    \centering
    \includegraphics[width=0.35\linewidth]{autofigures/cd_sweep_single.pdf}%
    \includegraphics[width=0.65\linewidth]{autofigures/cd_labels_short.pdf}
    \vspace{-8mm}
    \caption{(\textbf{left}) Performance of the Ada-GVAE with different $k$ on MPI3D. The algorithm adapts well to the unknown $k$ and benefits from sparser changes. (\textbf{center} and \textbf{right}) Comparison of Ada-ML-VAE with the vanilla ML-VAE which assumes group knowledge. We note that group knowledge may improve performance (\textbf{center}) but can also hurt when it is incomplete (\textbf{right}).}
    \label{fig:cd_sweep_labels}
    \vspace{-4mm}
\end{figure}

In Figure~\ref{fig:cd_sweep_labels} (left), we report the performance of Ada-GVAE without model selection for different values of $k$ on MPI3D (see Figure~\ref{fig:cd_sweep} in the appendix for the other data sets). We observe that Ada-GVAE is indeed adaptive to different values of $k$ and it achieves better performance when the change between the factors of variation is sparser. Note that our method is agnostic to the sharing pattern between the image pairs. In  applications where the number of shared factors is known to be constant, the performance may thus be further improved by injecting this knowledge into the inference procedure.

\looseness=-1\textbf{Summary} Our approach makes no assumption of which and how many factors are shared and successfully adapts to different values of $k$. The sparser the difference on the factors of variation, the more effective our method is in using weak supervision and learning disentangled representations.

\begin{figure*}[t]
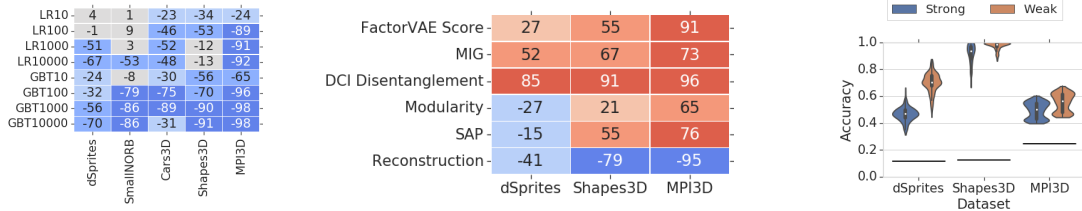

    \centering
    \quad \begin{minipage}{.25\textwidth}
    \centering\includegraphics[width=0.8\linewidth]{autofigures/disentanglement_vs_downstream_groupVAE.pdf}
    \end{minipage}%
    \quad
    \begin{minipage}{.4\textwidth}
    \includegraphics[width=0.8\linewidth]{autofigures/disentanglement_vs_downstream_strong.pdf}
    \end{minipage}%
    \begin{minipage}{.3\textwidth}
    \includegraphics[width=0.65\linewidth]{autofigures/generalization_gap.pdf}
    \end{minipage}
    \vspace{-3mm}
    \caption{(\textbf{left}) Rank correlation between our weakly-supervised reconstruction loss and performance of downstream prediction tasks with logistic regression (LR) and gradient boosted decision-trees (GBT) at different sample sizes for Ada-GVAE. We observe a general negative correlation that indicates that models with a low weakly-supervised reconstruction loss may also be more accurate. (\textbf{center}) Rank correlation between the strong generalization accuracy under covariate shifts and disentanglement scores as well as weakly-supervised reconstruction loss, for Ada-GVAE. (\textbf{right}) Distribution of vanilla (weak) generalization and under covariate shifts (strong generalization) for Ada-GVAE. The horizontal line corresponds to the accuracy of a naive classifier based on the prior only.}
    \label{fig:downstream}
    \vspace{-3mm}
\end{figure*}

\subsection{Supervision-performance trade-offs}\label{sec:results_group}
\looseness=-1The case $k=1$ where we actually know which factor of variation is not shared was previously considered in~\cite{bouchacourt2017multi,gvae2019,shu2020weakly}. Clearly, this additional knowledge should lead to improvements over our method. On the other hand, this information may be correct but incomplete in practice: For every pair of images, we know about one factor of variation that has changed but it may not be the only one. We therefore also consider the setup where $k=\texttt{Rnd}$ but the algorithm is only informed about one factor. Note that the original GVAE assumes group knowledge, so we directly compare its performance with our Ada-GVAE. We defer the comparison with ML-VAE~\cite{bouchacourt2017multi} and with the GAN-based approaches of~\cite{shu2020weakly} to Appendix~\ref{sec:results_group_app}.

In Figure~\ref{fig:cd_sweep_labels} (center and right), we observe that when $k=1$, the knowledge of which factor was changed generally improves the performance of weakly-supervised methods on MPI3D. On the other hand, the GVAE is not robust to incomplete knowledge as its performance degrades when the factor that is labeled as non-shared is not the only one. The performance degradation is stronger on the data sets with more factors of variation (dSprites/Shapes3D/MPI3D) as can be seen in Figure~\ref{fig:cd_labels} in the appendix. This may not come as a surprise as group-based disentanglement methods all assume that the group knowledge is precise.

\looseness=-1\textbf{Summary} Whenever the groups are fully and precisely known, this information can be used to improve disentanglement. Even though our adaptive method does not use group annotations, its performance is often comparable to the methods of~\cite{bouchacourt2017multi,gvae2019,shu2020weakly}. On the other hand, in practical applications there may not be precise control of which factors have changed. In this scenario, relying on incomplete group knowledge significantly harms the performance of GVAE and ML-VAE as they assume exact group knowledge. A blend between our adaptive variant and the vanilla GVAE may further improve performance when only partial group knowledge is available.

\subsection{Are weakly-supervised representations useful?}\label{sec:results_downstream}
In this section, we investigate whether the representations learned by our Ada-GVAE are useful on a variety of tasks. 
We show that representations with small weakly-supervised reconstruction loss (the sum of the first two terms in \eqref{eq:mod-elbo}) achieve improved downstream performance~\cite{locatello2018challenging,locatello2019disentangling}, improved downstream generalization~\cite{PetJanSch17} under covariate shifts~\cite{shimodaira2000improving,quionero2009dataset,ben2010impossibility}, fairer downstream predictions~\cite{locatello2019fairness}, and improved sample complexity on an abstract reasoning task~\cite{van2019disentangled}.  
To the best of our knowledge, strong generalization under covariate shift has not been tested on disentangled representations before. 

\looseness=-1\textbf{Key insight} We remark that the usefulness insights of~\citet{locatello2018challenging,locatello2019disentangling,locatello2019fairness,van2019disentangled} are based on the assumption that disentangled representations can be learned without observing the factors of variation. They consider models trained without supervision and argue that \textit{some} of the \textit{supervised disentanglement scores} (which require explicit labeling of the factors of variation) correlate well with desirable properties. \emph{In stark contrast, we here show that all these properties can be achieved simultaneously using only weakly-supervised data.}

\subsubsection{Downstream performance}
In this section, we consider the prediction task of \citet{locatello2018challenging} that predicts the values of the factors of variation from the representation. We also evaluate whether our weakly-supervised reconstruction loss is a good proxy for downstream performance.
We use a setup identical to~\citet{locatello2018challenging} and train the same logistic regression and gradient boosted decision trees (GBT) on the learned representations using different sample sizes (\num{10}/\num{100}/\num{1000}/\num{10000}). All test sets contain \num{5000} examples.

\looseness=-1In Figure~\ref{fig:downstream} (left), we observe that the weakly-supervised reconstruction loss of Ada-GVAE is generally anti-correlated with downstream performance. The best weakly-supervised disentanglement methods thus learn representations that are useful for training accurate classifiers downstream.

\textbf{Summary} The weakly-supervised reconstruction loss of our Ada-GVAE is a useful proxy for downstream accuracy.

\begin{figure*}[t]
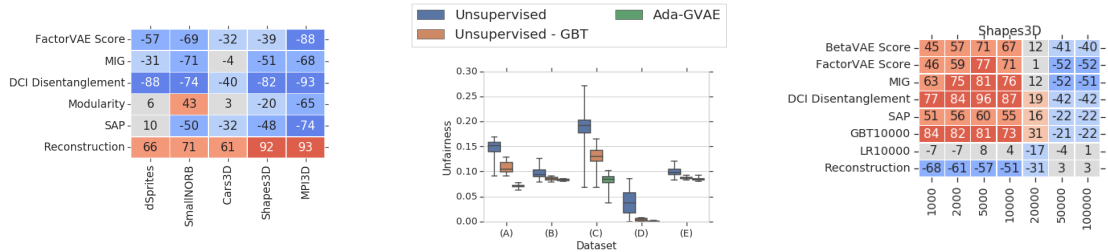

    \centering
    \begin{minipage}{.25\textwidth}
    \includegraphics[width=\linewidth]{autofigures/unfairness_disentanglement_rank.pdf}
    \end{minipage}%
    \qquad\quad
    \begin{minipage}{.25\textwidth}
    \includegraphics[width=\linewidth]{autofigures/Unfairness.pdf}
    \end{minipage}%
    \qquad
    \begin{minipage}{.25\textwidth}
    \includegraphics[width=\linewidth]{autofigures/rank_scores_vs_steps_number.pdf}
    \end{minipage}
    \vspace{-0.2cm}
    \caption{\looseness=-1(\textbf{left}) Rank correlation between both disentanglement scores and our weakly-supervised reconstruction loss with the unfairness of GBT10000 on all the data sets for Ada-GVAE. (\textbf{center}) Unfairness of the unsupervised methods with the semi-supervised model selection heuristic of~\cite{locatello2019fairness} and our weakly-supervised Ada-GVAE with $k=1$.
    (\textbf{right}) Rank correlation with down-stream accuracy of the abstract visual
    reasoning models of~\cite{van2019disentangled} throughout training (i.e., for different sample sizes).
    }\vspace{-0.3cm}
    \label{fig:fairness_abstract}
\end{figure*}

\subsubsection{Generalization under covariate shift}
Assume we have access to a large pool of unlabeled paired data and our goal is to solve a prediction task for which we have a smaller labeled training set. Both the labeled training set and test set are biased,  but with different biases. For example, we want to predict object shape but our training set contains \textit{only} red objects, whereas the test set does not contain \textit{any} red objects.
We create a biased training set by performing an intervention on a random factor of variation (other than the target variable), so that its value is constant in the whole training set. We perform another intervention on the test set, so that the same factor can take all other values. We train a GBT classifier on 10000 examples from the representations learned by Ada-GVAE. For each target factor of variation, we repeat the training of the classifier 10 times for different random interventions. For this experiment, we consider only dSprites, Shapes3D and MPI3D since Cars3D and SmallNORB are too small (after an intervention on their most fine grained factor of variation, they only contain 96 and 270 images respectively).

\looseness=-1In Figure~\ref{fig:downstream} (center) we plot the rank correlation between disentanglement scores and weakly-supervised reconstruction, and the results for generalization under covariate shifts for Ada-GVAE. We note that both the disentanglement scores and our weakly-supervised reconstruction loss are correlated with strong generalization. In Figure~\ref{fig:downstream} (right), we highlight the gap between the performance of a classifier trained on a normal train/test split (which we refer to as \textit{weak} generalization) as opposed to this covariate shift setting. We do not perform model selection, so we can show the performance of the whole range of representations.
We observe that there is a gap between weak and strong generalization but the distributions of accuracies significantly overlap and are significantly better than a naive classifier based on the prior distribution of the classes.

\textbf{Summary} Our results provide compelling evidence that disentanglement is useful for strong generalization under covariate shifts. The best Ada-GVAE models in terms of weakly-supervised reconstruction loss are useful for training classifiers that generalize under covariate shifts.

\subsubsection{Fairness}
\looseness=-1Recently, \citet{locatello2019fairness} showed that disentangled representations may be useful to train robust classifiers that are fairer to unobserved sensitive variables independent of the target variable. While they observed a strong correlation between demographic parity~\cite{calders2009building,zliobaite2015relation} and disentanglement, the applicability of their approach is limited by the fact that disentangled representations are difficult to identify without access to explicit observations of the factors of variation~\cite{locatello2018challenging}.

Our experimental setup is identical to the one of \citet{locatello2019fairness} and we measure \textit{unfairness} of a classifier as in \citet[Section 4]{locatello2019fairness}.
In Figure~\ref{fig:fairness_abstract} (left), we show that the weakly-supervised reconstruction loss of our Ada-GVAE correlates with unfairness as strongly as the disentanglement scores, even though the former can be computed without observing the factors of variation. In particular, we can perform model selection without observing the sensitive variable. In Figure~\ref{fig:fairness_abstract} (center), we show that our Ada-GVAE with $k=1$ and model selection allows us to train and identify fairer models compared to the unsupervised models of~\citet{locatello2019fairness}. Furthermore, their model selection heuristic is based on downstream performance which requires knowledge of the sensitive variable. From both plots we conclude that our weakly-supervised reconstruction loss is a good proxy for unfairness and allows us to train fairer classifiers in the setup of~\citet{locatello2019fairness} even if the sensitive variable is not observed.

\textbf{Summary} We showed that using weak supervision, we can train and identify fairer classifiers in the sense of demographic parity~\cite{calders2009building,zliobaite2015relation}. As opposed to~\citet{locatello2019fairness}, we do not need to observe the target variable and yet, our principled weakly-supervised approach outperforms their semi-supervised heuristic. 

\subsubsection{Abstract visual reasoning} 
\looseness=-1Finally, we consider the abstract visual reasoning task of~\citet{van2019disentangled}. This task is based on Raven's progressive matrices~\cite{raven1941standardization} and requires completing the bottom right missing panel of a sequence of context panels arranged in a $3\times 3$ grid (see Figure~\ref{fig:abstract_reasoning_app} (left) in the appendix). The algorithm is presented with six potential answers and needs to choose the correct one. To solve this task, the model has to infer the abstract relationships between the panels.
We replicate the experiment of~\citet{van2019disentangled} on Shapes3D under the same exact experimental conditions (see Appendix~\ref{sec:details} for more details).

\looseness=-1In Figure~\ref{fig:fairness_abstract} (right), one can see that at low sample sizes, the weakly-supervised reconstruction loss is strongly anti-correlated with performance on the abstract visual reasoning task. As previously observed by~\citet{van2019disentangled}, this benefit only occurs at low sample sizes.

\textbf{Summary} We demonstrated that training a relational network on the representations learned by our Ada-GVAE improves its sample efficiency. This result is in line with the findings of~\citet{van2019disentangled} where disentanglement was found to correlate positively with improved sample complexity.

\section{Conclusion}
\looseness=-1In this paper, we considered the problem of learning disentangled representations from pairs of non-i.i.d. observations sharing an unknown, random subset of factors of variation. We demonstrated that, under certain technical assumptions, the associated disentangled  generative model is identifiable. 
We extensively discussed the impact of the different supervision modalities, such as the degree of group-level supervision, and studied the impact of the (unknown) number of shared factors. These insights will be particularly useful to practitioners having access to specific domain knowledge.
Importantly, we show how to select models with strong performance on a diverse suite of downstream tasks \emph{without using supervised disentanglement metrics}, relying exclusively on weak supervision. This result is of great importance as the community is becoming increasingly interested in the practical benefits of disentangled representations~\cite{van2019disentangled,locatello2019fairness,creager2019flexibly,chao2019hybrid,iten2020discovering,chartsias2019disentangled,higgins2017darla}. 
Future work should apply the proposed framework to challenging real-world data sets where the factors of variation are not observed and extend it to an interactive setup involving reinforcement learning.

\paragraph{Acknowledgments:}
The authors thank Stefan Bauer, Ilya Tolstikhin, Sarah Strauss and Josip Djolonga
for helpful discussions and comments.
Francesco Locatello is supported by the Max Planck ETH Center for Learning Systems, by an ETH core grant (to Gunnar R\"atsch), and by a Google Ph.D. Fellowship.
This work was partially done while Francesco Locatello was at Google Research, Brain Team, Zurich. 
\bibliography{main}

\begin{thebibliography}{77}
\providecommand{\natexlab}[1]{#1}
\providecommand{\url}[1]{\texttt{#1}}
\expandafter\ifx\csname urlstyle\endcsname\relax
  \providecommand{\doi}[1]{doi: #1}\else
  \providecommand{\doi}{doi: \begingroup \urlstyle{rm}\Url}\fi

\bibitem[Adel et~al.(2018)Adel, Ghahramani, and Weller]{adel2018discovering}
Adel, T., Ghahramani, Z., and Weller, A.
\newblock Discovering interpretable representations for both deep generative
  and discriminative models.
\newblock In \emph{International Conference on Machine Learning}, pp.\  50--59,
  2018.

\bibitem[Ben-David et~al.(2010)Ben-David, Lu, Luu, and
  P{\'a}l]{ben2010impossibility}
Ben-David, S., Lu, T., Luu, T., and P{\'a}l, D.
\newblock Impossibility theorems for domain adaptation.
\newblock In \emph{International Conference on Artificial Intelligence and
  Statistics}, pp.\  129--136, 2010.

\bibitem[Bengio(2017)]{bengio2017consciousness}
Bengio, Y.
\newblock The consciousness prior.
\newblock \emph{arXiv:1709.08568}, 2017.

\bibitem[Bengio et~al.(2013)Bengio, Courville, and
  Vincent]{bengio2013representation}
Bengio, Y., Courville, A., and Vincent, P.
\newblock Representation learning: A review and new perspectives.
\newblock \emph{IEEE Transactions on Pattern Analysis and Machine
  Intelligence}, 35\penalty0 (8):\penalty0 1798--1828, 2013.

\bibitem[Bengio et~al.(2019)Bengio, Deleu, Rahaman, Ke, Lachapelle, Bilaniuk,
  Goyal, and Pal]{bengio2019meta}
Bengio, Y., Deleu, T., Rahaman, N., Ke, R., Lachapelle, S., Bilaniuk, O.,
  Goyal, A., and Pal, C.
\newblock A meta-transfer objective for learning to disentangle causal
  mechanisms.
\newblock \emph{arXiv:1901.10912}, 2019.

\bibitem[Bouchacourt et~al.(2018)Bouchacourt, Tomioka, and
  Nowozin]{bouchacourt2017multi}
Bouchacourt, D., Tomioka, R., and Nowozin, S.
\newblock Multi-level variational autoencoder: Learning disentangled
  representations from grouped observations.
\newblock In \emph{AAAI Conference on Artificial Intelligence}, 2018.

\bibitem[Burgess et~al.(2018)Burgess, Higgins, Pal, Matthey, Watters,
  Desjardins, and Lerchner]{burgess2018understanding}
Burgess, C.~P., Higgins, I., Pal, A., Matthey, L., Watters, N., Desjardins, G.,
  and Lerchner, A.
\newblock Understanding disentangling in beta-{VAE}.
\newblock \emph{arXiv:1804.03599}, 2018.

\bibitem[Calders et~al.(2009)Calders, Kamiran, and
  Pechenizkiy]{calders2009building}
Calders, T., Kamiran, F., and Pechenizkiy, M.
\newblock Building classifiers with independency constraints.
\newblock In \emph{IEEE International Conference on Data Mining Workshops},
  pp.\  13--18, 2009.

\bibitem[Chao et~al.(2019)Chao, Kulkarni, Goebel, and Fink]{chao2019hybrid}
Chao, M.~A., Kulkarni, C., Goebel, K., and Fink, O.
\newblock Hybrid deep fault detection and isolation: Combining deep neural
  networks and system performance models.
\newblock \emph{arXiv:1908.01529}, 2019.

\bibitem[Chartsias et~al.(2019)Chartsias, Joyce, Papanastasiou, Semple,
  Williams, Newby, Dharmakumar, and Tsaftaris]{chartsias2019disentangled}
Chartsias, A., Joyce, T., Papanastasiou, G., Semple, S., Williams, M., Newby,
  D.~E., Dharmakumar, R., and Tsaftaris, S.~A.
\newblock Disentangled representation learning in cardiac image analysis.
\newblock \emph{Medical Image Analysis}, 58:\penalty0 101535, 2019.

\bibitem[Chen \& Batmanghelich(2020)Chen and Batmanghelich]{chen2019weakly}
Chen, J. and Batmanghelich, K.
\newblock Weakly supervised disentanglement by pairwise similarities.
\newblock In \emph{AAAI Conference on Artificial Intelligence}, 2020.

\bibitem[Chen et~al.(2018)Chen, Li, Grosse, and Duvenaud]{chen2018isolating}
Chen, T.~Q., Li, X., Grosse, R., and Duvenaud, D.
\newblock Isolating sources of disentanglement in variational autoencoders.
\newblock In \emph{Advances in Neural Information Processing Systems}, 2018.

\bibitem[Cheung et~al.(2014)Cheung, Livezey, Bansal, and
  Olshausen]{cheung2014discovering}
Cheung, B., Livezey, J.~A., Bansal, A.~K., and Olshausen, B.~A.
\newblock Discovering hidden factors of variation in deep networks.
\newblock \emph{arXiv:1412.6583}, 2014.

\bibitem[Comon(1994)]{comon1994independent}
Comon, P.
\newblock Independent component analysis, a new concept?
\newblock \emph{Signal Processing}, 36\penalty0 (3):\penalty0 287--314, 1994.

\bibitem[Creager et~al.(2019)Creager, Madras, Jacobsen, Weis, Swersky, Pitassi,
  and Zemel]{creager2019flexibly}
Creager, E., Madras, D., Jacobsen, J.-H., Weis, M., Swersky, K., Pitassi, T.,
  and Zemel, R.
\newblock Flexibly fair representation learning by disentanglement.
\newblock In \emph{International Conference on Machine Learning}, pp.\
  1436--1445, 2019.

\bibitem[Dayan(1993)]{dayan1993improving}
Dayan, P.
\newblock Improving generalization for temporal difference learning: The
  successor representation.
\newblock \emph{Neural Computation}, 5\penalty0 (4):\penalty0 613--624, 1993.

\bibitem[Deng et~al.(2017)Deng, Navarathna, Carr, Mandt, Yue, Matthews, and
  Mori]{deng2017factorized}
Deng, Z., Navarathna, R., Carr, P., Mandt, S., Yue, Y., Matthews, I., and Mori,
  G.
\newblock Factorized variational autoencoders for modeling audience reactions
  to movies.
\newblock In \emph{IEEE Conference on Computer Vision and Pattern Recognition},
  2017.

\bibitem[Denton \& Birodkar(2017)Denton and Birodkar]{denton2017unsupervised}
Denton, E.~L. and Birodkar, V.
\newblock Unsupervised learning of disentangled representations from video.
\newblock In \emph{Advances in Neural Information Processing Systems}, 2017.

\bibitem[Duan et~al.(2019)Duan, Watters, Matthey, Burgess, Lerchner, and
  Higgins]{duan2019heuristic}
Duan, S., Watters, N., Matthey, L., Burgess, C.~P., Lerchner, A., and Higgins,
  I.
\newblock A heuristic for unsupervised model selection for variational
  disentangled representation learning.
\newblock \emph{arXiv:1905.12614}, 2019.

\bibitem[Eastwood \& Williams(2018)Eastwood and
  Williams]{eastwood2018framework}
Eastwood, C. and Williams, C.~K.
\newblock A framework for the quantitative evaluation of disentangled
  representations.
\newblock In \emph{International Conference on Learning Representations}, 2018.

\bibitem[F{\"o}ldi{\'a}k(1991)]{foldiak1991learning}
F{\"o}ldi{\'a}k, P.
\newblock Learning invariance from transformation sequences.
\newblock \emph{Neural Computation}, 3\penalty0 (2):\penalty0 194--200, 1991.

\bibitem[Fortuin et~al.(2019)Fortuin, H{\"u}ser, Locatello, Strathmann, and
  R{\"a}tsch]{fortuin2018deep}
Fortuin, V., H{\"u}ser, M., Locatello, F., Strathmann, H., and R{\"a}tsch, G.
\newblock Deep self-organization: Interpretable discrete representation
  learning on time series.
\newblock In \emph{International Conference on Learning Representations}, 2019.

\bibitem[Fraccaro et~al.(2017)Fraccaro, Kamronn, Paquet, and
  Winther]{fraccaro2017disentangled}
Fraccaro, M., Kamronn, S., Paquet, U., and Winther, O.
\newblock A disentangled recognition and nonlinear dynamics model for
  unsupervised learning.
\newblock In \emph{Advances in Neural Information Processing Systems}, 2017.

\bibitem[Gondal et~al.(2019)Gondal, W{\"u}thrich, Miladinovi{\'c}, Locatello,
  Breidt, Volchkov, Akpo, Bachem, Sch{\"o}lkopf, and Bauer]{gondal2019transfer}
Gondal, M.~W., W{\"u}thrich, M., Miladinovi{\'c}, D., Locatello, F., Breidt,
  M., Volchkov, V., Akpo, J., Bachem, O., Sch{\"o}lkopf, B., and Bauer, S.
\newblock On the transfer of inductive bias from simulation to the real world:
  a new disentanglement dataset.
\newblock In \emph{Advances in Neural Information Processing Systems}, 2019.

\bibitem[Goroshin et~al.(2015)Goroshin, Mathieu, and
  LeCun]{goroshin2015learning}
Goroshin, R., Mathieu, M.~F., and LeCun, Y.
\newblock Learning to linearize under uncertainty.
\newblock In \emph{Advances in Neural Information Processing Systems}, 2015.

\bibitem[Gresele et~al.(2019)Gresele, Rubenstein, Mehrjou, Locatello, and
  Sch\"olkopf]{gresele2019incomplete}
Gresele, L., Rubenstein, P.~K., Mehrjou, A., Locatello, F., and Sch\"olkopf, B.
\newblock The incomplete rosetta stone problem: Identifiability results for
  multi-view nonlinear ica.
\newblock In \emph{Conference on Uncertainty in Artificial Intelligence}, 2019.

\bibitem[Higgins et~al.(2017{\natexlab{a}})Higgins, Matthey, Pal, Burgess,
  Glorot, Botvinick, Mohamed, and Lerchner]{higgins2016beta}
Higgins, I., Matthey, L., Pal, A., Burgess, C., Glorot, X., Botvinick, M.,
  Mohamed, S., and Lerchner, A.
\newblock beta-{VAE}: Learning basic visual concepts with a constrained
  variational framework.
\newblock In \emph{International Conference on Learning Representations},
  2017{\natexlab{a}}.

\bibitem[Higgins et~al.(2017{\natexlab{b}})Higgins, Pal, Rusu, Matthey,
  Burgess, Pritzel, Botvinick, Blundell, and Lerchner]{higgins2017darla}
Higgins, I., Pal, A., Rusu, A., Matthey, L., Burgess, C., Pritzel, A.,
  Botvinick, M., Blundell, C., and Lerchner, A.
\newblock Darla: Improving zero-shot transfer in reinforcement learning.
\newblock In \emph{International Conference on Machine Learning},
  2017{\natexlab{b}}.

\bibitem[Higgins et~al.(2018)Higgins, Amos, Pfau, Racaniere, Matthey, Rezende,
  and Lerchner]{higgins2018towards}
Higgins, I., Amos, D., Pfau, D., Racaniere, S., Matthey, L., Rezende, D., and
  Lerchner, A.
\newblock Towards a definition of disentangled representations.
\newblock \emph{arXiv:1812.02230}, 2018.

\bibitem[Hochreiter \& Schmidhuber(1999)Hochreiter and
  Schmidhuber]{hochreiter1999feature}
Hochreiter, S. and Schmidhuber, J.
\newblock Feature extraction through lococode.
\newblock \emph{Neural Computation}, 11\penalty0 (3):\penalty0 679--714, 1999.

\bibitem[Hosoya(2019)]{gvae2019}
Hosoya, H.
\newblock Group-based learning of disentangled representations with
  generalizability for novel contents.
\newblock In \emph{International Joint Conference on Artificial Intelligence},
  pp.\  2506--2513, 2019.

\bibitem[Hsieh et~al.(2018)Hsieh, Liu, Huang, Fei-Fei, and
  Niebles]{hsieh2018learning}
Hsieh, J.-T., Liu, B., Huang, D.-A., Fei-Fei, L.~F., and Niebles, J.~C.
\newblock Learning to decompose and disentangle representations for video
  prediction.
\newblock In \emph{Advances in Neural Information Processing Systems}, 2018.

\bibitem[Hsu et~al.(2017)Hsu, Zhang, and Glass]{hsu2017unsupervised}
Hsu, W.-N., Zhang, Y., and Glass, J.
\newblock Unsupervised learning of disentangled and interpretable
  representations from sequential data.
\newblock In \emph{Advances in Neural Information Processing Systems}, 2017.

\bibitem[Huang et~al.(2017)Huang, Zhang, Zhang, Sanchez-Romero, Glymour, and
  Sch{\"o}lkopf]{HuaZhaZhaSanGlySch17}
Huang, B., Zhang, K., Zhang, J., Sanchez-Romero, R., Glymour, C., and
  Sch{\"o}lkopf, B.
\newblock Behind distribution shift: Mining driving forces of changes and
  causal arrows.
\newblock In \emph{{IEEE} International Conference on Data Mining}, pp.\
  913--918, 2017.

\bibitem[Hyvarinen \& Morioka(2016)Hyvarinen and
  Morioka]{hyvarinen2016unsupervised}
Hyvarinen, A. and Morioka, H.
\newblock Unsupervised feature extraction by time-contrastive learning and
  nonlinear ica.
\newblock In \emph{Advances in Neural Information Processing Systems}, 2016.

\bibitem[Hyvarinen \& Morioka(2017)Hyvarinen and
  Morioka]{hyvarinen2017nonlinear}
Hyvarinen, A. and Morioka, H.
\newblock Nonlinear ica of temporally dependent stationary sources.
\newblock In \emph{Artificial Intelligence and Statistics}, pp.\  460--469,
  2017.

\bibitem[Hyv{\"a}rinen \& Pajunen(1999)Hyv{\"a}rinen and
  Pajunen]{hyvarinen1999nonlinear}
Hyv{\"a}rinen, A. and Pajunen, P.
\newblock Nonlinear independent component analysis: Existence and uniqueness
  results.
\newblock \emph{Neural Networks}, 1999.

\bibitem[Hyvarinen et~al.(2019)Hyvarinen, Sasaki, and
  Turner]{hyvarinen2018nonlinear}
Hyvarinen, A., Sasaki, H., and Turner, R.~E.
\newblock Nonlinear ica using auxiliary variables and generalized contrastive
  learning.
\newblock In \emph{International Conference on Artificial Intelligence and
  Statistics}, 2019.

\bibitem[Iten et~al.(2020)Iten, Metger, Wilming, Del~Rio, and
  Renner]{iten2020discovering}
Iten, R., Metger, T., Wilming, H., Del~Rio, L., and Renner, R.
\newblock Discovering physical concepts with neural networks.
\newblock \emph{Physical Review Letters}, 124\penalty0 (1):\penalty0 010508,
  2020.

\bibitem[Karaletsos et~al.(2015)Karaletsos, Belongie, and
  R{\"a}tsch]{karaletsos2015bayesian}
Karaletsos, T., Belongie, S., and R{\"a}tsch, G.
\newblock Bayesian representation learning with oracle constraints.
\newblock \emph{arXiv:1506.05011}, 2015.

\bibitem[Ketchen \& Shook(1996)Ketchen and Shook]{ketchen1996application}
Ketchen, D.~J. and Shook, C.~L.
\newblock The application of cluster analysis in strategic management research:
  an analysis and critique.
\newblock \emph{Strategic Management Journal}, 17\penalty0 (6):\penalty0
  441--458, 1996.

\bibitem[Khemakhem et~al.(2019)Khemakhem, Kingma, and
  Hyv{\"a}rinen]{khemakhem2019variational}
Khemakhem, I., Kingma, D.~P., and Hyv{\"a}rinen, A.
\newblock Variational autoencoders and nonlinear {ICA}: A unifying framework.
\newblock \emph{arXiv:1907.04809}, 2019.

\bibitem[Kim \& Mnih(2018)Kim and Mnih]{kim2018disentangling}
Kim, H. and Mnih, A.
\newblock Disentangling by factorising.
\newblock In \emph{International Conference on Machine Learning}, 2018.

\bibitem[Kulkarni et~al.(2015)Kulkarni, Whitney, Kohli, and
  Tenenbaum]{kulkarni2015deep}
Kulkarni, T.~D., Whitney, W.~F., Kohli, P., and Tenenbaum, J.
\newblock Deep convolutional inverse graphics network.
\newblock In \emph{Advances in Neural Information Processing Systems}, 2015.

\bibitem[Kumar et~al.(2018)Kumar, Sattigeri, and
  Balakrishnan]{kumar2017variational}
Kumar, A., Sattigeri, P., and Balakrishnan, A.
\newblock Variational inference of disentangled latent concepts from unlabeled
  observations.
\newblock In \emph{International Conference on Learning Representations}, 2018.

\bibitem[LeCun et~al.(2004)LeCun, Huang, and Bottou]{lecun2004learning}
LeCun, Y., Huang, F.~J., and Bottou, L.
\newblock Learning methods for generic object recognition with invariance to
  pose and lighting.
\newblock In \emph{IEEE Conference on Computer Vision and Pattern Recognition},
  2004.

\bibitem[Locatello et~al.(2018)Locatello, Vincent, Tolstikhin, R{\"a}tsch,
  Gelly, and Sch{\"o}lkopf]{locatello2018clustering}
Locatello, F., Vincent, D., Tolstikhin, I., R{\"a}tsch, G., Gelly, S., and
  Sch{\"o}lkopf, B.
\newblock Competitive training of mixtures of independent deep generative
  models.
\newblock In \emph{Workshop at the 6th International Conference on Learning
  Representations (ICLR)}, 2018.

\bibitem[Locatello et~al.(2019{\natexlab{a}})Locatello, Abbati, Rainforth,
  Bauer, Sch{\"o}lkopf, and Bachem]{locatello2019fairness}
Locatello, F., Abbati, G., Rainforth, T., Bauer, S., Sch{\"o}lkopf, B., and
  Bachem, O.
\newblock On the fairness of disentangled representations.
\newblock In \emph{Advances in Neural Information Processing Systems},
  2019{\natexlab{a}}.

\bibitem[Locatello et~al.(2019{\natexlab{b}})Locatello, Bauer, Lucic, Gelly,
  Sch{\"o}lkopf, and Bachem]{locatello2018challenging}
Locatello, F., Bauer, S., Lucic, M., Gelly, S., Sch{\"o}lkopf, B., and Bachem,
  O.
\newblock Challenging common assumptions in the unsupervised learning of
  disentangled representations.
\newblock In \emph{International Conference on Machine Learning},
  2019{\natexlab{b}}.

\bibitem[Locatello et~al.(2020)Locatello, Tschannen, Bauer, R{\"a}tsch,
  Sch{\"o}lkopf, and Bachem]{locatello2019disentangling}
Locatello, F., Tschannen, M., Bauer, S., R{\"a}tsch, G., Sch{\"o}lkopf, B., and
  Bachem, O.
\newblock Disentangling factors of variation using few labels.
\newblock \emph{International Conference on Learning Representations}, 2020.

\bibitem[Mathieu et~al.(2016)Mathieu, Zhao, Ramesh, Sprechmann, and
  LeCun]{mathieu2016disentangling}
Mathieu, M.~F., Zhao, J.~J., Ramesh, A., Sprechmann, P., and LeCun, Y.
\newblock Disentangling factors of variation in deep representation using
  adversarial training.
\newblock In \emph{Advances in Neural Information Processing Systems}, 2016.

\bibitem[Narayanaswamy et~al.(2017)Narayanaswamy, Paige, Van~de Meent,
  Desmaison, Goodman, Kohli, Wood, and Torr]{narayanaswamy2017learning}
Narayanaswamy, S., Paige, T.~B., Van~de Meent, J.-W., Desmaison, A., Goodman,
  N., Kohli, P., Wood, F., and Torr, P.
\newblock Learning disentangled representations with semi-supervised deep
  generative models.
\newblock In \emph{Advances in Neural Information Processing Systems}, 2017.

\bibitem[Peters et~al.(2017)Peters, Janzing, and Sch{\"o}lkopf]{PetJanSch17}
Peters, J., Janzing, D., and Sch{\"o}lkopf, B.
\newblock \emph{Elements of Causal Inference - Foundations and Learning
  Algorithms}.
\newblock Adaptive Computation and Machine Learning Series. MIT Press, 2017.

\bibitem[Quionero-Candela et~al.(2009)Quionero-Candela, Sugiyama, Schwaighofer,
  and Lawrence]{quionero2009dataset}
Quionero-Candela, J., Sugiyama, M., Schwaighofer, A., and Lawrence, N.~D.
\newblock \emph{Dataset shift in machine learning}.
\newblock The MIT Press, 2009.

\bibitem[Raven(1941)]{raven1941standardization}
Raven, J.~C.
\newblock Standardization of progressive matrices, 1938.
\newblock \emph{British Journal of Medical Psychology}, 19\penalty0
  (1):\penalty0 137--150, 1941.

\bibitem[Reed et~al.(2014)Reed, Sohn, Zhang, and Lee]{reed2014learning}
Reed, S., Sohn, K., Zhang, Y., and Lee, H.
\newblock Learning to disentangle factors of variation with manifold
  interaction.
\newblock In \emph{International Conference on Machine Learning}, 2014.

\bibitem[Reed et~al.(2015)Reed, Zhang, Zhang, and Lee]{reed2015deep}
Reed, S., Zhang, Y., Zhang, Y., and Lee, H.
\newblock Deep visual analogy-making.
\newblock In \emph{Advances in Neural Information Processing Systems}, 2015.

\bibitem[Ridgeway(2016)]{ridgeway2016survey}
Ridgeway, K.
\newblock A survey of inductive biases for factorial representation-learning.
\newblock \emph{arXiv:1612.05299}, 2016.

\bibitem[Ridgeway \& Mozer(2018)Ridgeway and Mozer]{ridgeway2018learning}
Ridgeway, K. and Mozer, M.~C.
\newblock Learning deep disentangled embeddings with the f-statistic loss.
\newblock In \emph{Advances in Neural Information Processing Systems}, 2018.

\bibitem[Santoro et~al.(2018)Santoro, Hill, Barrett, Morcos, and
  Lillicrap]{santoro2018measuring}
Santoro, A., Hill, F., Barrett, D., Morcos, A., and Lillicrap, T.
\newblock Measuring abstract reasoning in neural networks.
\newblock In \emph{International Conference on Machine Learning}, pp.\
  4477--4486, 2018.

\bibitem[Schmidt et~al.(2007)Schmidt, Niculescu-Mizil, Murphy,
  et~al.]{schmidt2007learning}
Schmidt, M., Niculescu-Mizil, A., Murphy, K., et~al.
\newblock Learning graphical model structure using l1-regularization paths.
\newblock In \emph{AAAI}, volume~7, pp.\  1278--1283, 2007.

\bibitem[Sch{\"o}lkopf(2019)]{1911.10500}
Sch{\"o}lkopf, B.
\newblock Causality for machine learning, 2019.
\newblock arXiv:1911.10500.

\bibitem[Sch{\"o}lkopf et~al.(2012)Sch{\"o}lkopf, Janzing, Peters, Sgouritsa,
  Zhang, and Mooij]{scholkopf2012causal}
Sch{\"o}lkopf, B., Janzing, D., Peters, J., Sgouritsa, E., Zhang, K., and
  Mooij, J.
\newblock On causal and anticausal learning.
\newblock In \emph{International Conference on Machine Learning}, 2012.

\bibitem[Shimodaira(2000)]{shimodaira2000improving}
Shimodaira, H.
\newblock Improving predictive inference under covariate shift by weighting the
  log-likelihood function.
\newblock \emph{Journal of Statistical Planning and Inference}, 90\penalty0
  (2):\penalty0 227--244, 2000.

\bibitem[Shu et~al.(2020)Shu, Chen, Kumar, Ermon, and Poole]{shu2020weakly}
Shu, R., Chen, Y., Kumar, A., Ermon, S., and Poole, B.
\newblock Weakly supervised disentanglement with guarantees.
\newblock \emph{International Conference on Learning Representations}, 2020.

\bibitem[Sorrenson et~al.(2020)Sorrenson, Rother, and
  K{\"o}the]{sorrenson2020disentanglement}
Sorrenson, P., Rother, C., and K{\"o}the, U.
\newblock Disentanglement by nonlinear {ICA} with general incompressible-flow
  networks ({GIN}).
\newblock \emph{arXiv:2001.04872}, 2020.

\bibitem[Storck et~al.(1995)Storck, Hochreiter, and
  Schmidhuber]{storck1995reinforcement}
Storck, J., Hochreiter, S., and Schmidhuber, J.
\newblock Reinforcement driven information acquisition in non-deterministic
  environments.
\newblock In \emph{International Conference on Artificial Neural Networks},
  pp.\  159--164, 1995.

\bibitem[Suter et~al.(2019)Suter, Miladinovi{\'c}, Bauer, and
  Sch{\"o}lkopf]{suter2018interventional}
Suter, R., Miladinovi{\'c}, D., Bauer, S., and Sch{\"o}lkopf, B.
\newblock Interventional robustness of deep latent variable models.
\newblock In \emph{International Conference on Machine Learning}, 2019.

\bibitem[Thomas et~al.(2017)Thomas, Bengio, Fedus, Pondard, Beaudoin,
  Larochelle, Pineau, Precup, and Bengio]{thomas2018disentangling}
Thomas, V., Bengio, E., Fedus, W., Pondard, J., Beaudoin, P., Larochelle, H.,
  Pineau, J., Precup, D., and Bengio, Y.
\newblock Disentangling the independently controllable factors of variation by
  interacting with the world.
\newblock \emph{Learning Disentangled Representations Workshop at NeurIPS},
  2017.

\bibitem[Tschannen et~al.(2018)Tschannen, Bachem, and
  Lucic]{tschannen2018recent}
Tschannen, M., Bachem, O., and Lucic, M.
\newblock Recent advances in autoencoder-based representation learning.
\newblock \emph{arXiv:1812.05069}, 2018.

\bibitem[van Steenkiste et~al.(2019)van Steenkiste, Locatello, Schmidhuber, and
  Bachem]{van2019disentangled}
van Steenkiste, S., Locatello, F., Schmidhuber, J., and Bachem, O.
\newblock Are disentangled representations helpful for abstract visual
  reasoning?
\newblock In \emph{Advances in Neural Information Processing Systems}, 2019.

\bibitem[Whitney et~al.(2016)Whitney, Chang, Kulkarni, and
  Tenenbaum]{whitney2016understanding}
Whitney, W.~F., Chang, M., Kulkarni, T., and Tenenbaum, J.~B.
\newblock Understanding visual concepts with continuation learning.
\newblock \emph{arXiv:1602.06822}, 2016.

\bibitem[Yang et~al.(2015)Yang, Reed, Yang, and Lee]{yang2015weakly}
Yang, J., Reed, S.~E., Yang, M.-H., and Lee, H.
\newblock Weakly-supervised disentangling with recurrent transformations for
  3{D} view synthesis.
\newblock In \emph{Advances in Neural Information Processing Systems}, 2015.

\bibitem[Yingzhen \& Mandt(2018)Yingzhen and Mandt]{yingzhen2018disentangled}
Yingzhen, L. and Mandt, S.
\newblock Disentangled sequential autoencoder.
\newblock In \emph{International Conference on Machine Learning}, pp.\
  5656--5665, 2018.

\bibitem[Zhang et~al.(2015)Zhang, Gong, and
  Sch{\"o}lkopf]{zhang_multi-source_2015}
Zhang, K., Gong, M., and Sch{\"o}lkopf, B.
\newblock Multi-source domain adaptation: A causal view.
\newblock In \emph{AAAI Conference on Artificial Intelligence}, pp.\
  3150--3157, 2015.

\bibitem[Zhang et~al.(2017)Zhang, Huang, Zhang, Glymour, and
  Sch{\"o}lkopf]{Zhangetal17}
Zhang, K., Huang, B., Zhang, J., Glymour, C., and Sch{\"o}lkopf, B.
\newblock Causal discovery from nonstationary/heterogeneous data: Skeleton
  estimation and orientation determination.
\newblock In \emph{International Joint Conference on Artificial Intelligence},
  pp.\  1347--1353, 2017.

\bibitem[Zliobaite(2015)]{zliobaite2015relation}
Zliobaite, I.
\newblock On the relation between accuracy and fairness in binary
  classification.
\newblock \emph{arXiv:1505.05723}, 2015.

\end{thebibliography}
\bibliographystyle{icml2020}

\appendix
\onecolumn

\section{Proof of Theorem~\ref{thm:identifiability}}\label{sec:proof}
Recall that the true marginal likelihoods $p(\rvx_1|\cdot) = p(\rvx_2|\cdot)$, are completely specified through the smooth, invertible function $g^\star$. The corresponding posteriors $p(\cdot | \rvx_1) = p(\cdot | \rvx_2)$ are completely determined by $g^{\star-1}$. The model family for candidate marginal likelihoods $q(\rvx_1| \cdot) = q(\rvx_2| \cdot)$ and corresponding posteriors $q(\cdot | \rvx_1) = q(\cdot | \rvx_2)$ are hence conditional distributions specified by the set of smooth invertible functions $g \colon \gZ \to \gX$ and their inverses $g^{-1}$, respectively.

In order to prove identifiability, we show that every candidate posterior distribution $q(\hat \rvz| \rvx_1)$ (more precisely, the corresponding $g$) on the generative model~\eqref{eq:generative-model-1}--\eqref{eq:generative-model-4} satisfying the assumptions stated in Theorem~\ref{thm:identifiability} inverts $g^\star$ in the sense that the aggregate posterior $q(\hat \rvz) = \int q(\hat \rvz|\rvx_1) p(\rvx_1) \dx \rvx_1$ is a coordinate-wise reparameterization of $p(\rvz)$ up to permutation of the indices.
Crucially, while neither the latent variables nor the shared indices are directly observed, observing pairs of images allows us to verify whether a candidate distribution has the right factorization \eqref{eq:full-joint} and sharing structure imposed by $S$ or not.

The proof is composed of the following steps:
\begin{enumerate}
    \item We characterize the constraints that need to hold for the posterior $q(\hat \rvz|\rvx_1)$ (the associated $g^{-1}$) inverting $g^\star$ for fixed~$S$.
    \item We parameterize all candidate posteriors $q(\hat \rvz|\rvx_1)$ (the associated $g^{-1}$) as a function $g^\star$ for a fixed $S$.
    \item We show that, for fixed $S$, $q(\hat \rvz|\rvx_1)$ (the associated $g^{-1}$) has two disentangled coordinate subspaces, one corresponding to $S$ and one corresponding to $\bar S$, in the sense that varying $\rvz_S$ and keeping $\rvz_{\bar S}$ fixed results in changes of the coordinate subspace of $\hat \rvz$ corresponding to $S$ only, and vice versa.
    \item We show that randomly sampling $S$ implies that every candidate posterior has an aggregated posterior which is a coordinate-wise reparameterization of the distribution of the true factors of variation.
\end{enumerate}

\paragraph{Step 1} We start by noting that since any continuous distribution can be obtained from the standard uniform distribution (via the inverse cumulative distribution function), it is sufficient to simply set $p(\hat \rvz)$ to the $d$-dimensional standard uniform distribution and try to recover an axis-aligned, smooth, invertible function $g \colon \gZ \to \gX$ (which completely characterizes $q(\rvx_1 | \hat \rvz)$ and $q(\hat \rvz|\rvx_1)$ via its inverse) as well as the distribution $p(S)$.

Next, assume that $S$ is fixed but unknown, i.e., the following reasoning is conditionally on $S$. By the generative process \eqref{eq:generative-model-1}--\eqref{eq:generative-model-4} we know that all smooth, invertible candidate functions $g$ need to obey with probability $1$ (and irrespective of whether $p(\hat \rvz)$ or $p(\rvz)$ is used)
\begin{align}
    g^{-1}_i(\rvx_1) &= g^{-1}_i(\rvx_2) \quad \forall i \in T, \label{eq:gconstraint1v2}\\
    g^{-1}_j(\rvx_1) &\neq g^{-1}_j(\rvx_2) \quad \forall i,j \in \bar T, \label{eq:gconstraint2v2}
\end{align}
for all $(\rvx_1,\rvx_2) \in \supp(p(\rvx_1,\rvx_2|S))$, where $T \in \gS$ is arbitrary but fixed. $T$ indexes the the coordinate subspace in the image of $g^{-1}$ corresponding to the unknown coordinate subspace $S$ of shared factors of $\rvz$. Note that choosing $T \in \gS$ requires knowledge of $k$ ($d$ can be inferred from $p(\rvx_1, \rvx_2)$). Also note that $g^\star$ satisfies \eqref{eq:gconstraint1v2}--\eqref{eq:gconstraint2v2} for $T=S$. 

\paragraph{Step 2} All smooth, invertible candidate functions can be written as $g = g^\star \circ h$, where $h\colon [0,1]^d \to \gZ$ is a smooth invertible function with smooth inverse (using that the composition of smooth invertible functions is smooth and invertible) that maps the $d$-dimensional uniform distribution to $p(\rvz)$. 

We have $g^{-1} = h^{-1} \circ g^{\star-1}$ i.e., $g^{-1}(\vx_1) = h^{-1}(g^{\star-1}(\vx_1)) = h^{-1}(\rvz)$ and similarly $g^{-1}(\vx_2) = h^{-1}(f(\rvz, \tilde \rvz, S))$. Expressing now \eqref{eq:gconstraint1v2}--\eqref{eq:gconstraint2v2} through $h$ we have with probability $1$
\begin{align}
    h^{-1}_i(\rvz) &= h^{-1}_i(f(\rvz, \tilde \rvz, S)) \quad \forall i \in T, \label{eq:rconstraint1v2}\\
    h^{-1}_j(\rvz) &\neq h^{-1}_j(f(\rvz, \tilde \rvz, S)) \quad \forall i,j \in \bar T. \label{eq:rconstraint2v2}
\end{align}
Thanks to invertibility and smoothness of $h$ we know that $h^{-1} $ maps the coordinate subspace $S$ of $\gZ$ to a $(d-k)$-dimensional submanifold $\gM_S$ of $[0,1]^d$ and the coordinate subspace $\bar S$ to a $k$-dimensional sub-manifold $\gM_{\bar S}$ of $[0,1]^d$ that is disjoint from $\gM_S$. 

\paragraph{Step 3}
Next, we shall see that for a fixed $S$ the only admissible functions $h \colon [0,1]^d \rightarrow\sZ^d$ are identifying two groups of factors (corresponding to two orthogonal coordinate subspaces): Those in $S$ and those in $\bar S$. 

To see this, we prove that $h$ can only satisfy \eqref{eq:rconstraint1v2}--\eqref{eq:rconstraint2v2} if it aligns the coordinate subspace $S$ of $\gZ$ with the coordinate subspace $T$ of $[0,1]^d$ and $\bar S$ with $\bar T$. In other words, $\gM_S$ and $\gM_{\bar S}$ lie in the coordinate subspaces $T$ and $\bar T$, respectively, and the Jacobian of $h^{-1}$ is block diagonal with blocks of coordinates indexed by $T$ and $\bar T$.

By contradiction, if $\gM_{\bar S}$ does not lie in the coordinate subspace $\bar T$ then \eqref{eq:rconstraint1v2} is violated as $h$ is smooth and invertible but its arguments obey $\rvz_i \neq f(\rvz, \tilde \rvz, S)_i=\tilde \rvz_i$ for every $i \in \bar S$ with probability $1$.

Likewise, if $\gM_S$ does not lie in the coordinate subspace $T$ then \eqref{eq:rconstraint2v2} is violated as $h$ is smooth and invertible but its arguments satisfy $\rvz_{S} = f(\rvz, \tilde \rvz, S)_{S}$ with probability $1$. 

As a result, \eqref{eq:rconstraint1v2} and \eqref{eq:rconstraint2v2} can only be satisfied if $h^{-1}$ maps each coordinate in $S$ to a unique matching coordinate in $T$. In other words there exists a permutation $\pi$ on $[d]$ such that $h^{-1}$ can be simplified as $h^{-1} = \tilde h$, where
\begin{align}
    h^{-1}_T(\rvz) &= \tilde{h}_T(\rvz_{\pi(S)}) \label{eq:reparam-fixed-s-1v2} \\
    h^{-1}_{\bar T}(\rvz) &= \tilde{h}_{\bar T}(\rvz_{\pi(\bar S)}). \label{eq:reparam-fixed-s-2v2}
\end{align}
Note that the permutation is required because the choice of $T$ is arbitrary. This implies that the Jacobian of $\tilde{h}$ is block diagonal with blocks corresponding to coordinates indexed by $T$ and $\bar T$ (or equivalently $S$ and $\bar S$).

For fixed $S$, i.e., considering $p(\rvx_1,\rvx_2|S)$, we can recover the groups of factors in $g^\star_S$ and $g^\star_{\bar S}$ up to permutation of the factor indices. Note that this does not yet imply that we can recover all axis-aligned $g$ as the factors in $g_T$ and $g_{\bar T}$ may still be entangled with each other, i.e., $\tilde{h}$ is not axis aligned within $T$ and $\bar T$.

\paragraph{Step 4}
If now $S$ is drawn at random, we observe a mixture of distributions $p(\rvx_1,\rvx_2| S)$ (but not $S$ itself) and $g$ needs to associate every $(\rvx_1,\rvx_2) \in \supp(p(\rvx_1,\rvx_2|S))$ with one and only one $T$ to satisfy \eqref{eq:gconstraint1v2}--\eqref{eq:gconstraint2v2}, for every $S \in \supp(p(S))$.

Indeed, suppose that $(\rvx_1, \rvx_2)$ are distributed according to a mixture of $p(\rvx_1,\rvx_2| S=S_1)$ and $p(\rvx_1,\rvx_2| S=S_2)$ with $S_1, S_2 \in \supp(p(S)), S_1 \neq S_2$. Then \eqref{eq:gconstraint1v2} can only be satisfied with probability $1$ for a subset of coordinates of size $|S_1 \cap S_2| < d-k$ due to invertibility and smoothness of $g$, but $|T| = d-k$. The same reasoning applies for mixtures of more than two subsets of $p(\rvx_1,\rvx_2| S)$. Therefore, \eqref{eq:gconstraint1v2} cannot be satisfied for $(\rvx_1, \rvx_2)$ drawn from a mixture of distribution $p(\rvx_1,\rvx_2| S)$ but associated with a single $T$.

Conversely, for a given $S$, all $(\rvx_1,\rvx_2) \in \supp(p(\rvx_1,\rvx_2|S))$ need to be associated with the same $T$ due to invertibility and smoothness of $g$. in more detail, all $(\rvx_1,\rvx_2) \in \supp(p(\rvx_1,\rvx_2|S))$ will share the same $d-k$-dimensional coordinate subspace due to \eqref{eq:rconstraint1v2}--\eqref{eq:rconstraint2v2} and therefore cannot be associated with two different $T$ as $|T|=d-k$.

Further, note that due to the smoothness and invertibility of $g$, for every pair of associated $S_1, T_1$ and $S_2, T_2$ we have $|S_1 \cap S_2| = |T_1 \cap T_2|$ and $|S_1 \cup S_2| = |T_1 \cup T_2|$. The assumption
\begin{equation} \label{eq:ps-conditionv2}
    P(S \cap S' = \{i\}) > 0 \quad \forall i \in [d] \quad \text{and} \quad S, S' \sim p(S)
\end{equation}
hence implies that we ``observe'' every factor through $(\rvx_1, \rvx_2) \sim p(\rvx_1, \rvx_2)$ as the intersection of two sets $S_1,S_2$, and this intersection will be reflected as the intersection of the corresponding two coordinate subspaces $T_1, T_2$. This, together with \eqref{eq:reparam-fixed-s-1v2}--\eqref{eq:reparam-fixed-s-2v2} finally implies
\begin{align}
    h^{-1}_i(\rvz) &= \tilde{h}_i(z_{\pi(i)}) \quad \forall i \in [d]
\end{align}
for some permutation $\pi$ on $[d]$. This in turns imply that the Jacobian of $\tilde{h}$ is diagonal. 

Therefore, by change of variables formula we have
\begin{equation} \label{eq:reparam}
q(\hat\rvz) = p(\tilde{h}( \rvz_{\pi([d])})) \left|\mathrm{det}\frac{\partial}{\partial\rvz_{\pi([d])}}\tilde{h}\right| = \prod_{i=1}^d p(\tilde{h}_i( z_{\pi(i)})) \left|\frac{\partial}{\partial z_{\pi(i)}}\tilde{h}_i\right|
\end{equation}
where the second equality is a consequence of the Jacobian being diagonal, and $|\partial \tilde{h}_i / \partial z_{\pi(i)}| \neq 0, \forall i,$ thanks to $\tilde{h}\colon \gZ \to [0,1]^d$ being invertible on $\gZ$. From \eqref{eq:reparam}, we can see that $q(\hat\rvz)$ is a coordinate-wise reparameterization of $p(\rvz)$ up to permutation of the indices. As a consequence, a change in a coordinate of $\rvz$ implies a change in the unique corresponding coordinate of $\hat\rvz$, so $q(\hat \rvz|\rvx_1)$ (or, equivalently, $g$) disentangles the factors of variation.

\paragraph{Final remarks} The considered generative model is identifiable up to coordinate-wise reparametrization of the factors. $p(S)$ can then be recovered $p(\rvx_1, \rvx_2)$ via $g$. Note that \eqref{eq:ps-conditionv2} effectively ensures that to a weak supervision signal is available for each factor of variation.

\section{Implementation Details}~\label{sec:details}
We base our study on the \texttt{disentanglement\_lib} of~\cite{locatello2018challenging}. Here, we report for completeness all the hyperparameters used in our study. Our code will be released as part of the \texttt{disentanglement\_lib}.

In our study, fix the architecture (Table~\ref{table:architecture_main}) along with all other hyperparameters (Table~\ref{table:fixed_param_main}) except for one hyperparameter for each model (Table~\ref{table:sweep_main}). All hyperparameters for the unsupervised models are identical to~\cite{locatello2018challenging}. As our methods penalize the rate term in the ELBO similarly to $\beta$-VAE, we use the same hyperparameter range. We however note that in most cases, our model selection technique selects $\beta=1$. Exploring a different range for $\beta$ smaller than one is beyond the scope of this work. For the unsupervised methods we use the same \num{50} random seeds of~\cite{locatello2018challenging}. For the weakly-supervised methods, we use \num{10}.

\textbf{Downstream Task} The vanilla downstream task is based on~\cite{locatello2018challenging}. For each representation, we sample training sets of sizes $\num{10}$, $\num{100}$, $\num{1000}$ and $\num{10000}$. The test set always contains $\num{5000}$ points. The downstream task consists in predicting the value of each factor of variation from the representation. We use the same two models of~\cite{locatello2018challenging}: a cross validated logistic regression from Scikit-learn with 10 different values for the regularization strength ($Cs = 10$) and $5$ folds and a gradient boosting classifier (GBT) from Scikit-learn with default parameters.

\textbf{Downstream Task with Covariate Shift} We consider the same setup of the normal downstream task, but we only train a gradient boosted classifier with \num{10000} examples ($GBT10000$). For every target factor of variation we repeat 10 times the following process: sample another factor of variation uniformly and fix its value over the whole training set to an uniformly sampled value. The test set contains only examples where the intervened factors take values that are different from the one in the training set. We report the average test performance.

\textbf{Fairness Downstream Task} The fairness downstream task is based on~\cite{locatello2019fairness}. We train the same $GBT10000$ on each representation predicting each factor of variation and measure the unfairness using the formula in their Section~4.

\textbf{Abstract reasoning task}
We use the same Shapes3D simplified data set when training the relational network (scale and azimuth can only take four values instead of 8 and 16 to make the task feasible for humans). We consider the case where the rows in the grid have either 1, 2, or 3 constant ground-truth factors.
We train the same relational model~\cite{santoro2018measuring} as in~\cite{van2019disentangled} (with identical hyperparameters) on the frozen representations of our adaptive methods. 

We use hyperparameters identical to~\cite{van2019disentangled} which are reported here for completeness. The downstream classifier is the  \emph{Wild Relation Networks (WReN)} model of~\cite{santoro2018measuring}. 
For the experiments, we use the following random search space over the hyper-parameters. The optimizer's parameters are depicted in Table~\ref{table:wren}.
The edge MLP $g$ has either 256 or 512 hidden units and  2, 3, or 4 hidden layers.
The graph MLP $f$ has either 128 or 256 hidden units and 1 or 2 hidden layers before the final linear layer to compute the score. We also uniformly sample whether we apply no dropout, dropout of 0.25, dropout of 0.5, or dropout of 0.75 to units before this last layer and 10 random seeds.

\begin{table*}
\centering
\caption{Encoder and Decoder architecture for the main experiment.}
\vspace{2mm}
\begin{tabular}{l  l}
\toprule
\textbf{Encoder} & \textbf{Decoder}\\
\midrule 
Input: $64\times 64 \times$ number of channels & Input: $\R^{10}$\\
$4\times 4$ conv, 32 ReLU, stride 2 & FC, 256 ReLU\\
$4\times 4$ conv, 32 ReLU, stride 2 & FC, $4\times 4\times 64$ ReLU\\
$4\times 4$ conv, 64 ReLU, stride 2 & $4\times 4$ upconv, 64 ReLU, stride 2\\
$4\times 4$ conv, 64 ReLU, stride 2 & $4\times 4$ upconv, 32 ReLU, stride 2\\
FC 256, F2 $2\times 10$ & $4\times 4$ upconv, 32 ReLU, stride 2\\
& $4\times 4$ upconv, number of channels, stride 2\\
\bottomrule
\end{tabular}
\label{table:architecture_main}
\end{table*}

\begin{table*}
\centering
\caption{Model's hyperparameters. We allow a sweep over a single hyperparameter for each model.}
\vspace{2mm}
\begin{tabular}{l l  l}
\toprule
\textbf{Model} & \textbf{Parameter} & \textbf{Values}\\
\midrule 
$\beta$-VAE & $\beta$ & $[1,\ 2,\ 4,\ 6,\ 8,\ 16]$\\
AnnealedVAE & $c_{max}$ & $[5,\ 10,\ 25,\ 50,\ 75,\ 100]$\\
& iteration threshold & $100000$\\
& $\gamma$ & $1000$\\
FactorVAE & $\gamma$ & $[10,\ 20,\ 30,\ 40,\ 50,\ 100]$\\
DIP-VAE-I & $\lambda_{od}$ & $[1,\ 2,\ 5,\ 10,\ 20,\ 50]$\\
&$\lambda_{d}$ & $10\lambda_{od}$\\
DIP-VAE-II & $\lambda_{od}$ & $[1,\ 2,\ 5,\ 10,\ 20,\ 50]$\\
&$\lambda_{d}$ & $\lambda_{od}$\\
$\beta$-TCVAE & $\beta$ & $[1,\ 2,\ 4,\ 6,\ 8,\ 10]$\\
GVAE & $\beta$ & $[1,\ 2,\ 4,\ 6,\ 8,\ 16]$\\
Ada-GVAE & $\beta$ & $[1,\ 2,\ 4,\ 6,\ 8,\ 16]$\\
ML-VAE & $\beta$ & $[1,\ 2,\ 4,\ 6,\ 8,\ 16]$\\
Ada-ML-VAE & $\beta$ & $[1,\ 2,\ 4,\ 6,\ 8,\ 16]$\\

\bottomrule
\end{tabular}
\label{table:sweep_main}
\end{table*}

\begin{table}[t]\caption{Other fixed hyperparameters.}\label{table:fixed_param_main}
\centering
  \begin{subtable}[t]{0.3\linewidth}
    \centering
    \small
    \begin{tabular}{l  l}
      \toprule
      \textbf{Parameter} & \textbf{Values}\\
      \midrule 
      Batch size & $64$\\
      Latent space dimension & $10$\\
      Optimizer & Adam\\
      Adam: beta1 & 0.9\\
      Adam: beta2 & 0.999\\
      Adam: epsilon & 1e-8\\
      Adam: learning rate & 0.0001\\
      Decoder type & Bernoulli\\
      Training steps & 300000\\
      \bottomrule
    \end{tabular}
    \caption{Hyperparameters common to each of the considered methods.}
  \end{subtable}%
  \hspace{5mm}
  \begin{subtable}[t]{0.3\linewidth}
    \centering
    \begin{tabular}{l  }
      \toprule
      \textbf{Discriminator} \\
      \midrule 
      FC, 1000 leaky ReLU\\
      FC, 1000 leaky ReLU\\
      FC, 1000 leaky ReLU\\
      FC, 1000 leaky ReLU\\
      FC, 1000 leaky ReLU\\
      FC, 1000 leaky ReLU\\
      FC, 2 \\
      \bottomrule
    \end{tabular}
    \caption{Architecture for the discriminator in FactorVAE.}\label{table:discriminator_main}
  \end{subtable}%
  \hspace{5mm}
  \begin{subtable}[t]{.3\linewidth}
    \centering
    \begin{tabular}{l  l}
      \toprule
      \textbf{Parameter} & \textbf{Values}\\
      \midrule 
      Batch size & $64$\\
      Optimizer & Adam\\
      Adam: beta1 & 0.5\\
      Adam: beta2 & 0.9\\
      Adam: epsilon & 1e-8\\
      Adam: learning rate & 0.0001\\
      \bottomrule
    \end{tabular}
    \caption{Parameters for the discriminator in FactorVAE.}\label{table:param_discriminator_main}
  \end{subtable}
\end{table}

\begin{table*}
    \centering
   \begin{tabular}{l  l}
      \toprule
      \textbf{Parameter} & \textbf{Values}\\
      \midrule 
      Batch size & $32$\\
      Optimizer & Adam\\
      Adam: beta1 & 0.9\\
      Adam: beta2 & 0.999\\
      Adam: epsilon & 1e-8\\
      Adam: learning rate & $[0.01, 0.001, 0.0001]$\\
      \bottomrule
    \end{tabular}
    \caption{Parameters for the optimizer in the \emph{WReN}.}\label{table:wren}
\end{table*}
\clearpage
\section{Additional Results}\label{sec:results:app}
\subsection{Section~\ref{sec:results:learning_weak_sup}}~\label{sec:results:learning_weak_sup_app}
\begin{figure*}[t]
    \centering
    \includegraphics[width=\linewidth]{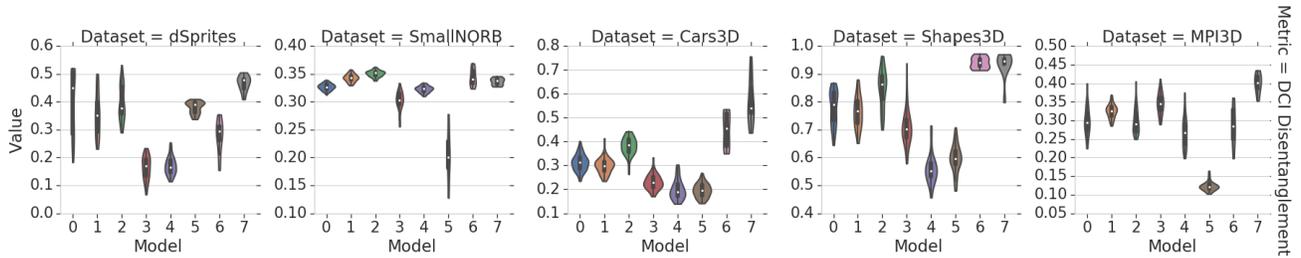}
    \caption{Our adaptive variants of the group-based disentanglement methods with \emph{weakly-supervised model selection} based on the reconstruction loss are competitive with \emph{fully supervised model selection} on the unsupervised models. In this experiment, we consider the case where the number of shared factors of variation is random and different for every pair.
    Legend: 0=$\beta$-VAE, 1=FactorVAE, 2=$\beta$-TCVAE, 3=DIP-VAE-I, 4=DIP-VAE-II, 5=AnnealedVAE, 6=Ada-ML-VAE, 7=Ada-GVAE}
    \label{fig:selection_app}
\end{figure*}

In Figure~\ref{fig:selection_app}, we show that our methods are competitive even with fully supervised model selection on the unsupervised methods.

\begin{figure*}[t]
    \centering
    \includegraphics[width=\linewidth]{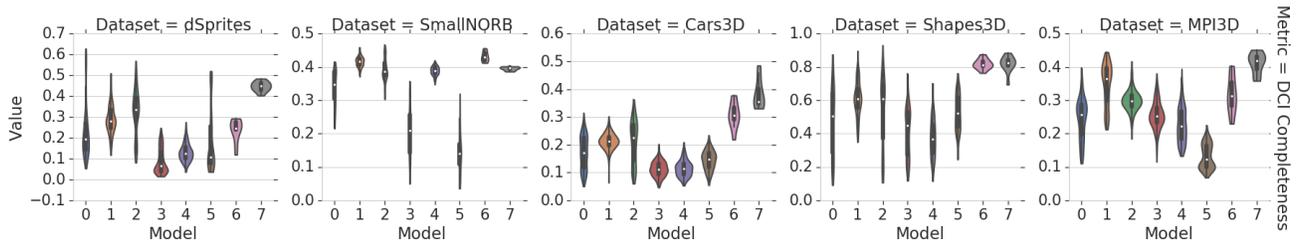}
    \caption{Our adaptive variants of the group-based disentanglement methods are competitive with unsupervised methods also in terms of Completeness. In this experiment, we consider the case where the number of shared factors of variation is random and different for every pair.
    Legend: 0=$\beta$-VAE, 1=FactorVAE, 2=$\beta$-TCVAE, 3=DIP-VAE-I, 4=DIP-VAE-II, 5=AnnealedVAE, 6=Ada-ML-VAE, 7=Ada-GVAE}
    \label{fig:selection_completeness}
\end{figure*}

\begin{figure*}[t]
    \centering
    \includegraphics[width=\linewidth]{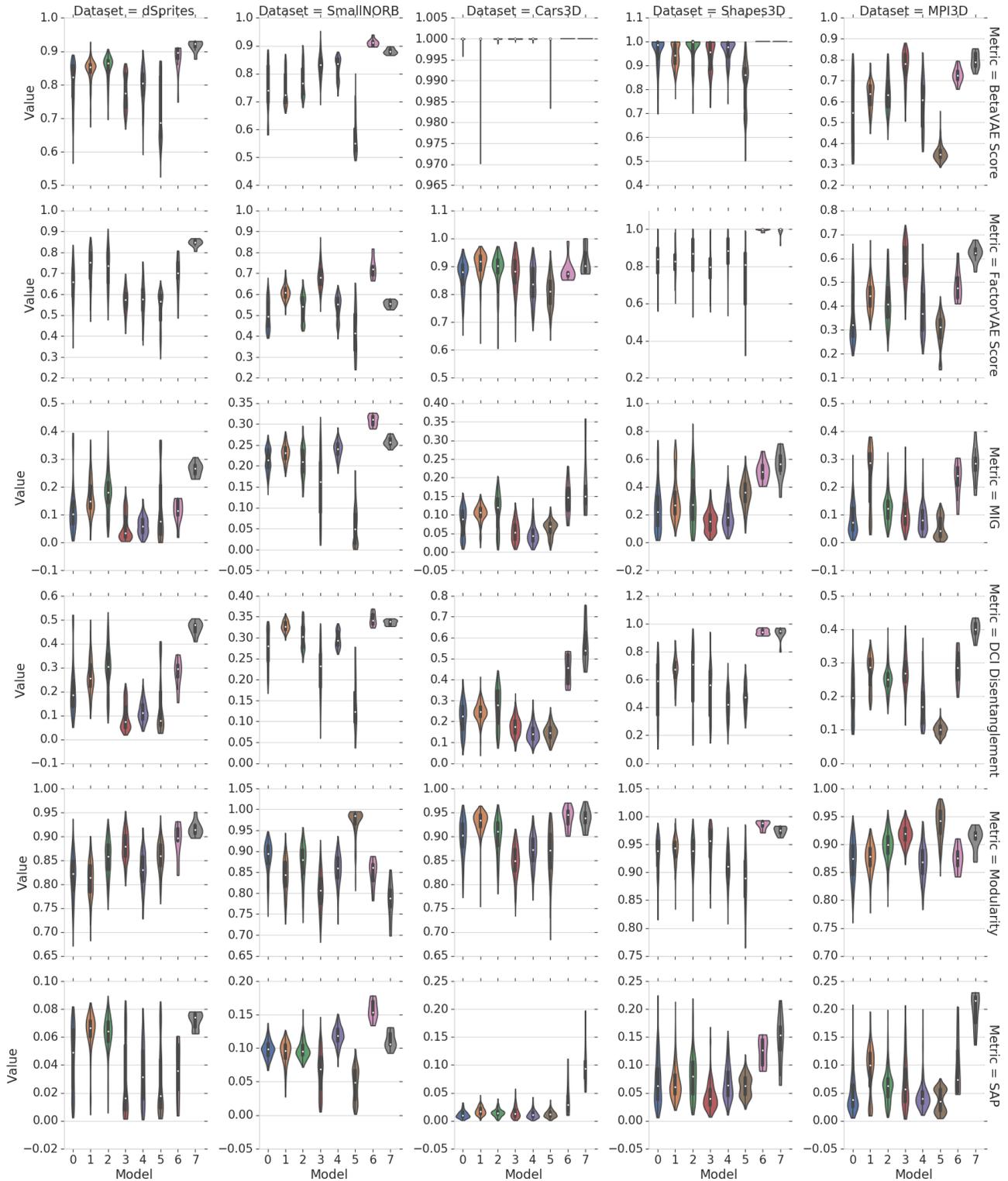}
    \caption{Our adaptive variants of the group-based disentanglement methods are competitive with unsupervised methods on all disentanglement scores. In this experiment, we consider the case where the number of shared factors of variation is random and different for every pair.
    Legend: 0=$\beta$-VAE, 1=FactorVAE, 2=$\beta$-TCVAE, 3=DIP-VAE-I, 4=DIP-VAE-II, 5=AnnealedVAE, 6=Ada-ML-VAE, 7=Ada-GVAE}
    \label{fig:selection_all}
\end{figure*}

\begin{figure*}[t]
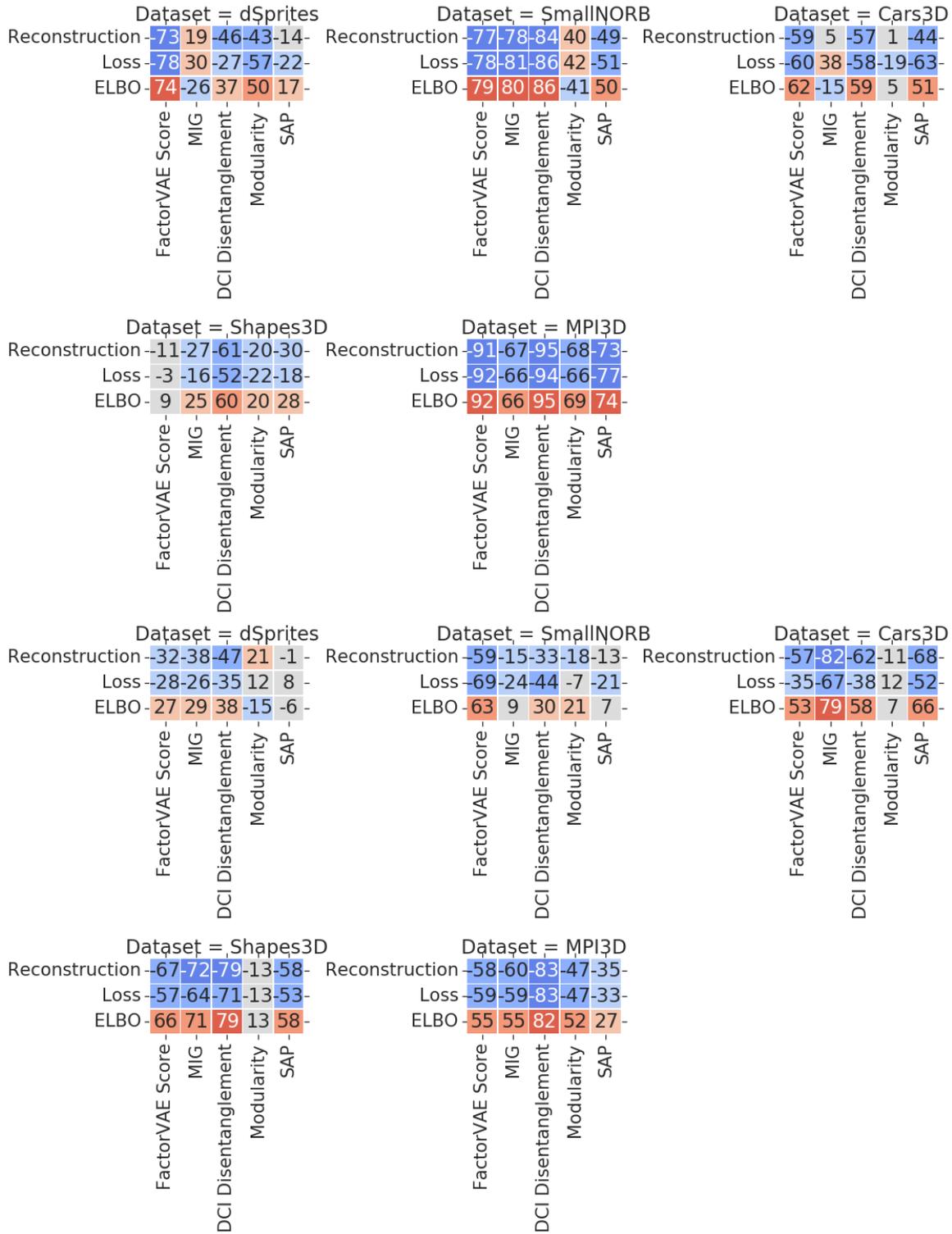

    \centering
    \includegraphics[width=0.9\linewidth]{autofigures/reconstriction_corr_common_dec.pdf}
    \includegraphics[width=0.9\linewidth]{autofigures/reconstriction_corr_mlvae.pdf}
    \caption{Rank correlation between training metrics and disentanglement scores for Ada-GVAE (top) and Ada-ML-VAE (bottom).}
    \label{fig:rank_corr}
\end{figure*}

While our main analysis is focused on DCI Disentanglement~\cite{eastwood2018framework}, we report in Figure~\ref{fig:selection_all} the performance of out methods when evaluated using each disentanglement score as well as Completeness~\cite{eastwood2018framework} in Figure~\ref{fig:selection_completeness}. The median values for all the models in Figure~\ref{fig:selection_all} are depicted in Tables~\ref{table:selection_all_ds}-~\ref{table:selection_all_mpi}. Overall, we observe that the trends we observed in Section~\ref{sec:results:learning_weak_sup} for DCI Disentanglement can be observed also for the other disentanglement scores (with the partial exception of Modularity~\cite{ridgeway2016survey}). In Figure~\ref{fig:rank_corr} we show that the disentanglement metrics are consistently correlated with the training metrics. We chose the weakly-supervised reconstruction loss for model selection but ELBO and overall Loss are also suitable.

\begin{table}\centering
\begin{tabular}{lrrrrrr}
\toprule
{} &  BetaVAE Score &  FactorVAE Score &   MIG &  DCI Disentanglement &  Modularity &  SAP \\
Model                  &                &                  &       &                      &             &      \\
\midrule
$\beta$-VAE            &          82.3\% &            66.0\% & 10.2\% &                18.6\% &       82.2\% & 4.9\% \\
FactorVAE              &          85.3\% &            75.0\% & 14.9\% &                25.6\% &       81.4\% & 6.7\% \\
$\beta$-TCVAE          &          86.4\% &            73.6\% & 18.0\% &                30.4\% &       85.8\% & 6.4\% \\
DIP-VAE-I              &          77.4\% &            57.2\% &  3.5\% &                 7.4\% &       87.9\% & 1.6\% \\
DIP-VAE-II             &          80.4\% &            57.6\% &  5.9\% &                11.0\% &       83.1\% & 3.1\% \\
AnnealedVAE            &          68.6\% &            56.5\% &  7.6\% &                 7.7\% &       86.0\% & 1.8\% \\
Ada-ML-VAE        &          89.6\% &            70.1\% & 11.5\% &                29.4\% &       89.7\% & 3.6\% \\
Ada-GVAE &          92.3\% &            84.7\% & 26.6\% &                47.9\% &       91.3\% & 7.4\% \\
\bottomrule
\end{tabular}
\caption{Median disentanglement scores on dSprites for the models in Figure~\ref{fig:selection_all}.}\label{table:selection_all_ds}
\end{table}
\begin{table}\centering
\begin{tabular}{lrrrrrr}
\toprule
{} &  BetaVAE Score &  FactorVAE Score &   MIG &  DCI Disentanglement &  Modularity &   SAP \\
Model                  &                &                  &       &                      &             &       \\
\midrule
$\beta$-VAE            &          74.0\% &            49.5\% & 21.4\% &                28.0\% &       89.5\% &  9.8\% \\
FactorVAE              &          72.4\% &            60.8\% & 23.2\% &                32.7\% &       84.4\% &  9.6\% \\
$\beta$-TCVAE          &          76.5\% &            54.2\% & 21.0\% &                30.2\% &       88.0\% &  9.6\% \\
DIP-VAE-I              &          83.1\% &            68.0\% & 16.2\% &                23.2\% &       80.6\% &  6.9\% \\
DIP-VAE-II             &          83.5\% &            55.1\% & 24.1\% &                29.3\% &       86.0\% & 11.8\% \\
AnnealedVAE            &          55.0\% &            41.3\% &  4.9\% &                12.3\% &       98.5\% &  4.9\% \\
Ada-ML-VAE        &          91.0\% &            72.1\% & 31.1\% &                34.1\% &       86.1\% & 15.3\% \\
Ada-GVAE &          87.9\% &            55.5\% & 25.6\% &                33.8\% &       78.8\% & 10.6\% \\
\bottomrule
\end{tabular}
\caption{Median disentanglement scores on SmallNORB for the models in Figure~\ref{fig:selection_all}.}\label{table:selection_all_SN}
\end{table}
\begin{table}\centering
\begin{tabular}{lrrrrrr}
\toprule
{} &  BetaVAE Score &  FactorVAE Score &   MIG &  DCI Disentanglement &  Modularity &  SAP \\
Model                  &                &                  &       &                      &             &      \\
\midrule
$\beta$-VAE            &         100.0\% &            87.9\% &  8.8\% &                22.5\% &       90.2\% & 1.0\% \\
FactorVAE              &         100.0\% &            91.8\% & 10.6\% &                24.5\% &       93.4\% & 1.7\% \\
$\beta$-TCVAE          &         100.0\% &            90.2\% & 12.0\% &                27.8\% &       91.0\% & 1.4\% \\
DIP-VAE-I              &         100.0\% &            88.2\% &  5.3\% &                17.4\% &       84.8\% & 1.2\% \\
DIP-VAE-II             &         100.0\% &            83.7\% &  4.3\% &                13.9\% &       87.2\% & 1.0\% \\
AnnealedVAE            &         100.0\% &            81.0\% &  6.8\% &                14.6\% &       87.1\% & 1.1\% \\
Ada-ML-VAE        &         100.0\% &            87.4\% & 14.7\% &                45.6\% &       94.6\% & 2.8\% \\
Ada-GVAE &         100.0\% &            90.2\% & 15.0\% &                54.0\% &       93.9\% & 9.4\% \\
\bottomrule
\end{tabular}
\caption{Median disentanglement scores on Cars3D for the models in Figure~\ref{fig:selection_all}.}\label{table:selection_all_c3}
\end{table}
\begin{table}\centering
\begin{tabular}{lrrrrrr}
\toprule
{} &  BetaVAE Score &  FactorVAE Score &   MIG &  DCI Disentanglement &  Modularity &   SAP \\
Model                  &                &                  &       &                      &             &       \\
\midrule
$\beta$-VAE            &          98.6\% &            83.9\% & 22.0\% &                58.8\% &       93.8\% &  6.2\% \\
FactorVAE              &          94.2\% &            82.5\% & 27.0\% &                67.2\% &       94.3\% &  6.1\% \\
$\beta$-TCVAE          &          99.8\% &            86.8\% & 27.1\% &                70.9\% &       93.8\% &  7.9\% \\
DIP-VAE-I              &          95.6\% &            79.7\% & 15.2\% &                55.9\% &       95.6\% &  4.0\% \\
DIP-VAE-II             &          97.8\% &            88.4\% & 18.1\% &                41.9\% &       91.0\% &  6.3\% \\
AnnealedVAE            &          86.1\% &            80.9\% & 35.9\% &                47.4\% &       89.0\% &  6.2\% \\
Ada-ML-VAE        &         100.0\% &           100.0\% & 50.9\% &                94.0\% &       98.8\% & 12.7\% \\
Ada-GVAE &         100.0\% &           100.0\% & 56.2\% &                94.6\% &       97.5\% & 15.3\% \\
\bottomrule
\end{tabular}
\caption{Median disentanglement scores on Shapes3D for the models in Figure~\ref{fig:selection_all}.}\label{table:selection_all_s3}
\end{table}
\begin{table}\centering
\begin{tabular}{lrrrrrr}
\toprule
{} &  BetaVAE Score &  FactorVAE Score &   MIG &  DCI Disentanglement &  Modularity &   SAP \\
Model                  &                &                  &       &                      &             &       \\
\midrule
$\beta$-VAE            &          54.6\% &            32.2\% &  7.2\% &                19.5\% &       87.4\% &  3.7\% \\
FactorVAE              &          63.8\% &            44.3\% & 28.6\% &                28.7\% &       87.8\% &  9.9\% \\
$\beta$-TCVAE          &          63.1\% &            40.9\% & 12.1\% &                25.0\% &       89.9\% &  6.2\% \\
DIP-VAE-I              &          78.1\% &            57.7\% &  9.6\% &                26.8\% &       91.9\% &  5.7\% \\
DIP-VAE-II             &          60.6\% &            36.9\% &  8.1\% &                16.9\% &       86.8\% &  4.0\% \\
AnnealedVAE            &          34.6\% &            31.3\% &  4.3\% &                10.1\% &       94.2\% &  3.5\% \\
Ada-ML-VAE        &          72.6\% &            47.6\% & 24.1\% &                28.5\% &       87.5\% &  7.4\% \\
Ada-GVAE &          78.9\% &            62.1\% & 28.4\% &                40.1\% &       91.6\% & 21.5\% \\
\bottomrule
\end{tabular}
\caption{Median disentanglement scores on MPI3D for the models in Figure~\ref{fig:selection_all}.}\label{table:selection_all_mpi}
\end{table}

\clearpage
\subsection{Section~\ref{sec:results_sup}}~\label{sec:results_sup_app}
\begin{figure*}[t]
    \centering
    \includegraphics[width=\linewidth]{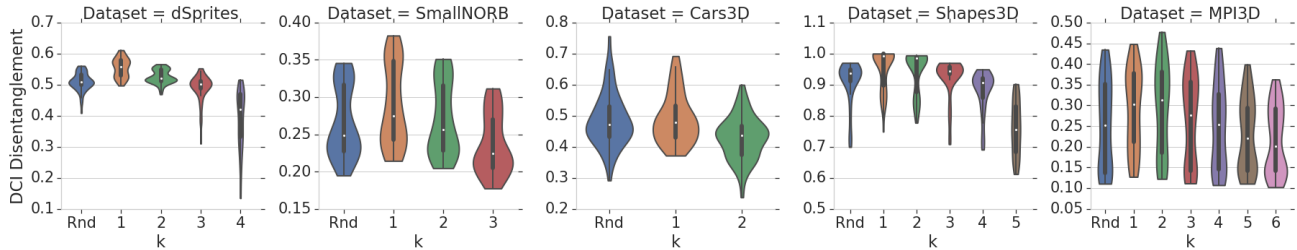}
    \caption{Performance of the Ada-GVAE with different degrees of supervision in the data. The best performances are when $k=1$---only one factor is changed in each pair---and they consistently degrade the fewer factors are shared until only a single factor of variation is shared. In the most general case, each pair has a different number of shared factors and the performance is consistent with the trend observed before.}
    \label{fig:cd_sweep}
\end{figure*}
\begin{figure*}[t]
    \centering
    \includegraphics[width=\linewidth]{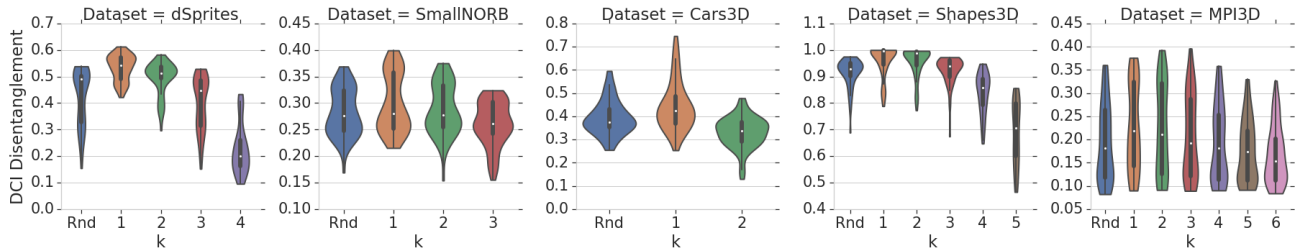}
    \caption{Performance of the Ada-ML-VAE with different amounts of supervision in the data. The best performances are when $k=1$ -- only one factor is changed -- and they consistently degrade the fewer factors are shared until only a single factor of variation is shared. In the most general case, each pair has a different amount of shared factors and the performance are consistent with the trend observed before.}
    \label{fig:mlvae_sweep}
\end{figure*}

Performance of Ada-GVAE~\ref{fig:cd_sweep} and Ada-ML-VAE~\ref{fig:mlvae_sweep} for different values of $k$. Generally, we observe that performances are best when the change between the pictures is sparser, i.e., $k=1$. We again note that the higher is $k$ the more similar the performances are with the vanilla $\beta$-VAE. 

\subsection{Section~\ref{sec:results_group}}~\label{sec:results_group_app}

\begin{figure*}[t]
    \centering
    \includegraphics[width=\linewidth]{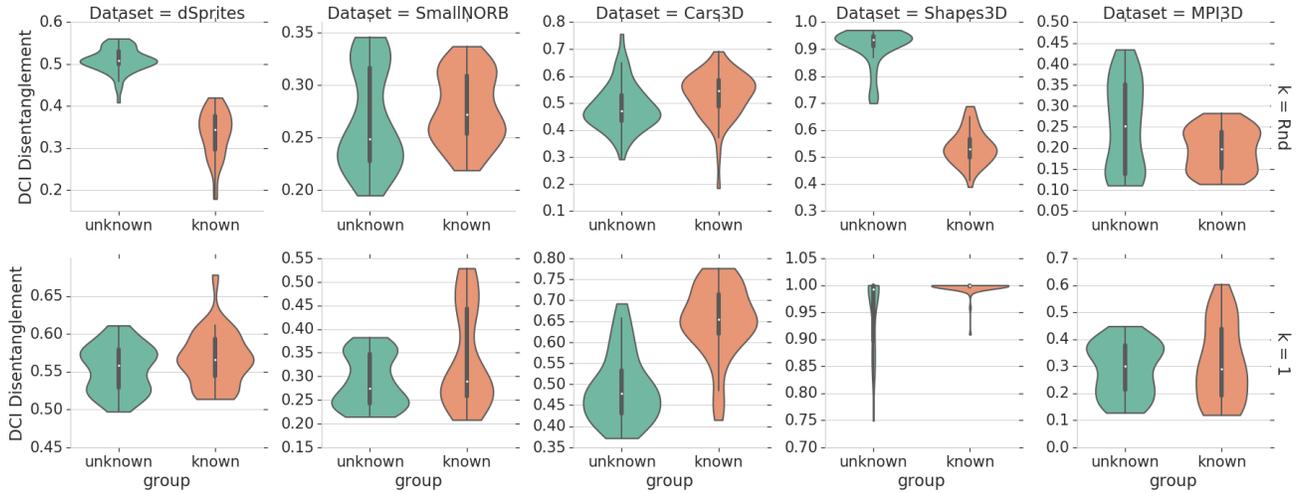}
    \caption{Comparison of Ada-GVAE with the vanilla GVAE which requires group knowledge. We note that group knowledge can improve disentanglement but can also significantly hurt when it is incomplete. Top row: $k=\texttt{Rnd}$, bottom row: $k=1$. }
    \label{fig:cd_labels}
\end{figure*}

\begin{figure*}[t]
    \centering
    \includegraphics[width=\linewidth]{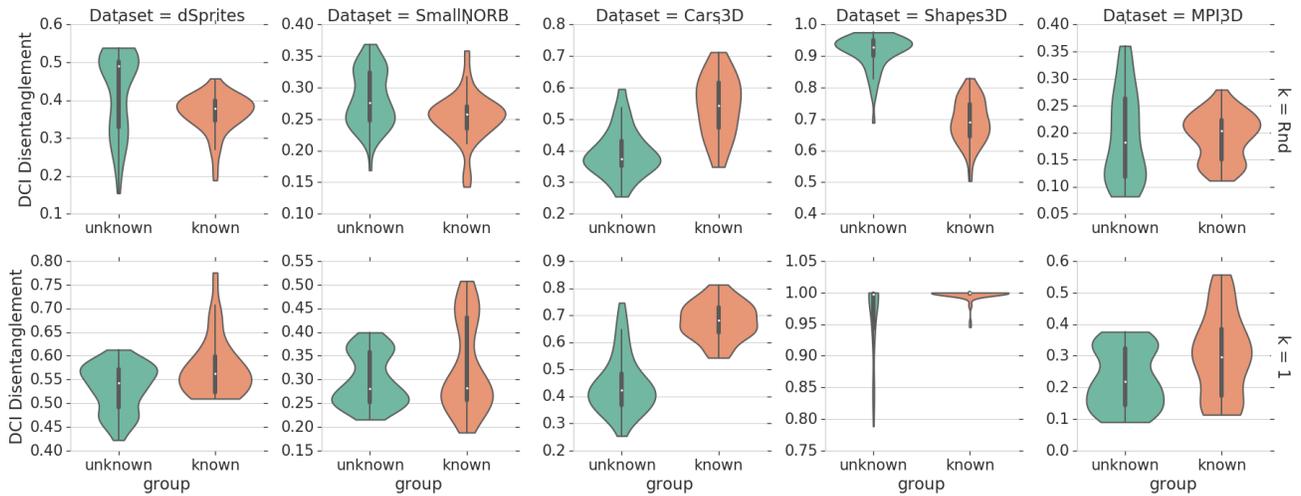}
    \caption{Comparison of Ada-ML-VAE with the vanilla ML-VAE which assumes group knowledge. We note that group knowledge improves performances but can also significantly hurt when it is incomplete.}
    \label{fig:mlvae_labels}
\end{figure*}

\begin{figure*}[t]
    \centering
    \includegraphics[width=\linewidth]{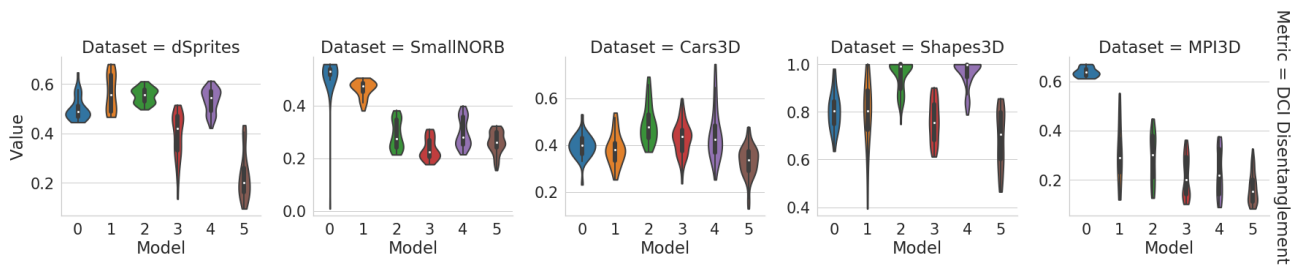}
    \caption{Comparison between the Change and Share GAN-based approach of~\cite{shu2020weakly} without model selection. Legend 0=Change, 1=Share, 2=Ada-GVAE $k=1$, 3=Ada-GVAE $k=d-1$, 4=Ada-ML-VAE $k=1$, 5=Ada-ML-VAE $k=d-1$. We remark that these methods are not directly comparable as (1) the experimental conditions are different and (2)~\citet{shu2020weakly} have access to additional supervision (which factor is shared or changed).}
    \label{fig:vae_vs_gan}
\end{figure*}

In Figures~\ref{fig:cd_labels} and~\ref{fig:mlvae_labels}, we observe that, regardless of the averaging, when $k=1$ and the different factor is known to the algorithm, this knowledge improves the disentanglement. However, when this knowledge is incomplete it harms the disentanglement. In Figure~\ref{fig:vae_vs_gan} we show how our method compare with the \textit{Change} and \textit{Share} GAN-based approaches of~\cite{shu2020weakly}. The goal of this plot is to show that ball-park the two approaches achieves similar results. We stress that strong conclusions should not be drawn from this plot as~\cite{shu2020weakly} used different experimental conditions from ours. Finally, we remark that~\cite{shu2020weakly} assume access to which factors was either shared or changed in the pair. Our method was designed to benefit from very similar images and without any additional annotation, so it is not completely surprising that when $k=d-1$ our performances are worse. It is however interesting to notice how the GAN based methods perform especially well on the data sets SmallNORB and MPI3D where VAE based approaches struggle with reconstruction as the objects are either too detailed or too small.


\clearpage
\subsection{Section~\ref{sec:results_downstream}}~\label{sec:results_downstream_app}
\begin{figure*}[t]
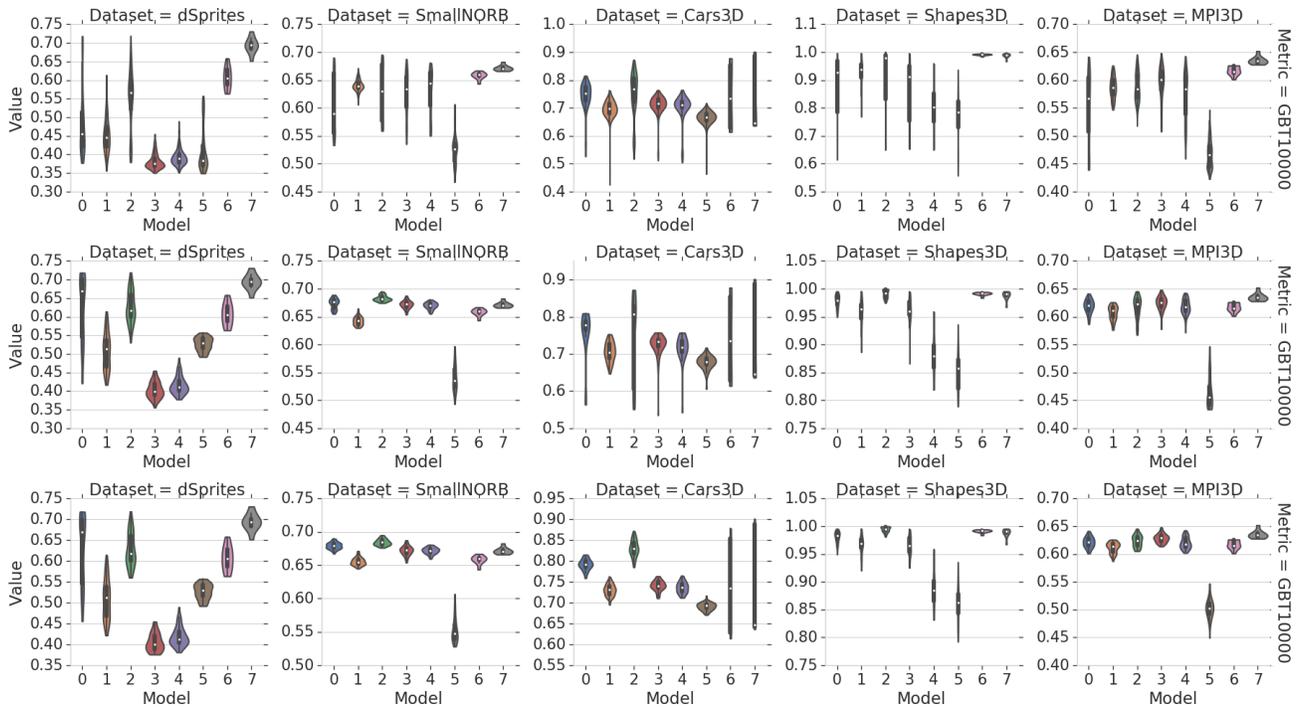

    \centering
    \includegraphics[width=\linewidth]{autofigures/no_selection_GBT.pdf}
    \includegraphics[width=\linewidth]{autofigures/dci_selection_GBT.pdf}
    \includegraphics[width=\linewidth]{autofigures/selection_GBT.pdf}
    \caption{Our adaptive variants of the group-based disentanglement methods are competitive with unsupervised methods in terms of Downstream performance. In this experiment, we consider the case where the number of shared factors of variation is random and different for every pair. We test different model selection techniques for the unsupervised methods: (top) no model selection, (middle) model selection with DCI Disentanglement and (bottom) model selection with test downstream performance.
    Legend: 0=$\beta$-VAE, 1=FactorVAE, 2=$\beta$-TCVAE, 3=DIP-VAE-I, 4=DIP-VAE-II, 5=AnnealedVAE, 6=Ada-ML-VAE, 7=Ada-GVAE}
    \label{fig:selection_all_downstream}
\end{figure*}

\begin{figure*}[t]
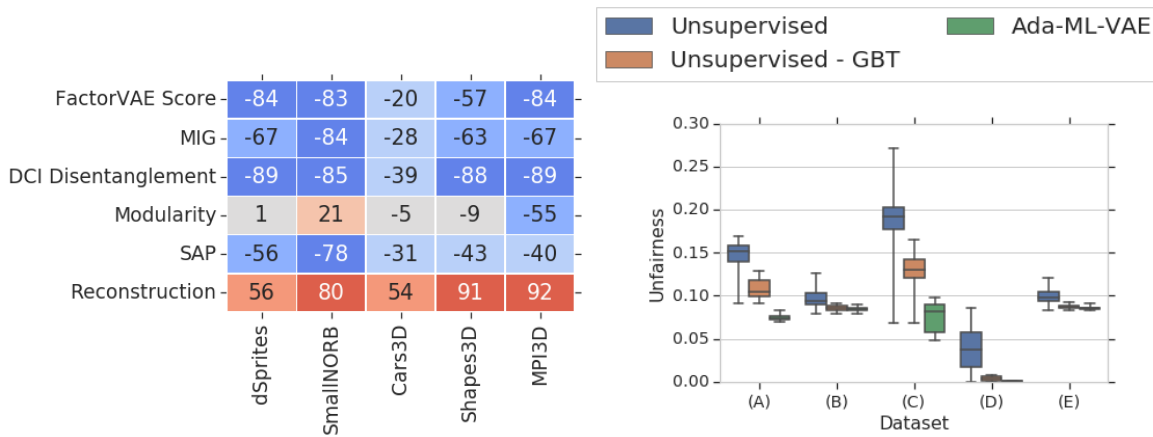

    \centering
    \includegraphics[width=0.45\linewidth]{autofigures/unfairness_disentanglement_rank_app.pdf}
    \includegraphics[width=0.45\linewidth]{autofigures/Unfairness_app.pdf}
    \caption{(\textbf{left}) Rank correlation between both disentanglement scores and the weakly-supervised reconstruction loss of our Ada-ML-VAE with the unfairness of GBT10000 on all the data sets. (\textbf{right}) Unfairness of the unsupervised methods with the semi-supervised model selection heuristic of~\cite{locatello2019fairness} and our Ada-ML-VAE with $k=1$. From both plots, we conclude that out weakly-supervised reconstruction loss is a good proxy for the unfairness and allows to train fairer classifiers in the setup of~\cite{locatello2019fairness} even if the sensitive variable is not observed.}
    \label{fig:fairness_app}
\end{figure*}

\begin{figure*}[t]
    \centering
    \begin{minipage}{.4\textwidth}
    \includegraphics[width=\linewidth]{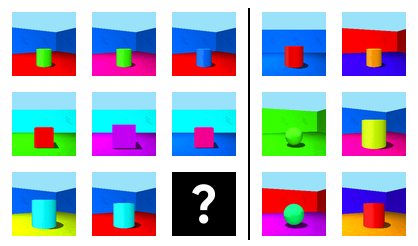}
    \end{minipage}%
    \begin{minipage}{.4\textwidth}
    \includegraphics[width=\linewidth]{autofigures/rank_scores_vs_steps_number_app.pdf}
    \end{minipage}
    \caption{(\textbf{left}) Example of the abstract visual reasoning task of~\cite{van2019disentangled}. The solution is  the panel in the central row on the right. (\textbf{right}) Rank correlation between disentanglement metrics, prediction accuracy, weakly-supervised reconstruction and down-stream accuracy of the abstract visual
reasoning models throughout training (i.e., for different sample sizes) for the Ada-ML-VAE.  }
    \label{fig:abstract_reasoning_app}
\end{figure*}

In Figure~\ref{fig:selection_all_downstream}, we show the performance of our approach in terms of downstream performance compared to the unsupervised methods (top) without model selection, (middle) performing model selection with the DCI Disentanglement score and (bottom) performing model selection on the test downstream performance. Our models are always selected based on their reconstruction error. We observe that our method is competitive in terms of downstream performance even if we allow model selection on the test score for the baselines. 
In Figure~\ref{fig:downstream_app}, we show the figure analogous to Figure~\ref{fig:downstream} for the Ada-ML-VAE. We observe that the trends are comparable to the ones we observed for the Ada-GVAE. In Figures~\ref{fig:fairness_app} and~\ref{fig:abstract_reasoning_app}, we show the results on the fairness and abstract reasoning downstream task for the Ada-ML-VAE. Overall, we observe that the conclusions we drew for the Ada-GVAE is valid for the Ada-ML-VAE too: good models in terms of weakly-supervised reconstruction loss are useful on all the considered downstream tasks.

\begin{figure*}[t]
    \centering
    \begin{minipage}{.3\textwidth}
    \includegraphics[width=\linewidth]{autofigures/disentanglement_vs_downstream_app.pdf}
    \end{minipage}%
    \begin{minipage}{.3\textwidth}
    \includegraphics[width=\linewidth]{autofigures/disentanglement_vs_downstream_strong_app.pdf}
    \end{minipage}%
    \begin{minipage}{.3\textwidth}
    \includegraphics[width=\linewidth]{autofigures/generalization_gap_app.pdf}
    \end{minipage}
    \caption{(\textbf{left}) Rank correlation between our weakly-supervised reconstruction loss and performance of downstream prediction tasks with Logistic Regression (LR) and Gradient Boosted decision-Trees at different sample sizes for the Ada-ML-VAE. We observe a general negative correlation that indicates that models with a good weakly-supervised reconstruction loss may also be more accurate. (\textbf{center}) Rank correlation between disentanglement scores and weakly-supervised reconstruction loss with strong generalization under covariate shifts for the Ada-ML-VAE. (\textbf{right}) Generalization gap between weak and strong generalization for the Ada-ML-VAE over all models. The horizontal line is the accuracy of random chance. }
    \label{fig:downstream_app}
\end{figure*}

\end{document}